%% file: main.tex
\pdfoutput=1

\documentclass[11pt]{article}
\usepackage[table,xcdraw]{xcolor}

\usepackage[preprint]{acl}

\usepackage{times}
\usepackage{latexsym}

\usepackage[T1]{fontenc}

\usepackage[utf8]{inputenc}

\usepackage{microtype}

\usepackage{inconsolata}

\usepackage[noautocite]{annotation}

\usepackage{float}
\usepackage{graphicx}
\usepackage{pifont}
\usepackage{wrapfig}
\usepackage{multirow}
\usepackage{tikz}
\usepackage{soul}
\usepackage{tcolorbox}
\usepackage{geometry}
\usepackage{svg}
\usepackage{amsmath}
\usepackage{soul}
\usepackage{makecell}
\usepackage{subfigure}
\usepackage[subfigure]{tocloft}
\usepackage{multirow}
\usepackage{bbold}
\usepackage{subcaption}
\usepackage{enumitem}
\usepackage{cleveref}
\usepackage{booktabs}
\usepackage{placeins}
\usepackage{makecell}
\usepackage{svg}
\usepackage{authblk}
\tcbuselibrary{breakable}
\usepackage{colortbl}

\newcommand{\judge}{\textbf{DRIFTJudge}}
\newcommand{\policy}{\textbf{DRIFTPolicy}}
\newcommand{\metric}{\textbf{FOCUS}}
\newcommand{\dataset}{\textbf{DRIFTEval}}

\title{Stay Focused: Problem Drift in Multi-Agent Debate}

\author{
 \textbf{Jonas Becker\textsuperscript{1,2,*}},
 \textbf{Lars Benedikt Kaesberg\textsuperscript{1}},
 \textbf{Andreas Stephan\textsuperscript{3}},\\
 \textbf{Jan Philip Wahle\textsuperscript{1}},
 \textbf{Terry Ruas\textsuperscript{1}},
 \textbf{Bela Gipp\textsuperscript{1}}
\\
 \textsuperscript{1}University of Göttingen, Germany; \textsuperscript{2}LKA NRW, Germany; \textsuperscript{3}University of Vienna, Austria
\\
 \small{
   \textbf{\textsuperscript{*}Correspondence:} \href{mailto:jonas.becker@uni-goettingen.de}{jonas.becker@uni-goettingen.de}
 }
}

\DeclareRobustCommand{\hlgreen}[1]{{\sethlcolor{green!50}\hl{#1}}}
\DeclareRobustCommand{\hlred}[1]{{\sethlcolor{red!50}\hl{#1}}}

\begin{document}
\maketitle
\AddAnnotationRef{}

\begin{abstract}
Multi-agent debate --- multiple instances of large language models discussing problems in turn-based interaction --- has shown promise for solving knowledge and reasoning tasks.
However, these methods show limitations when solving complex problems that require longer reasoning chains.
We analyze how multi-agent debate drifts away from the initial problem over multiple turns, thus harming task performance.
We define this phenomenon as \textbf{problem drift} and quantify its presence across ten tasks (i.e., three generative, three knowledge, three reasoning, and one instruction-following task).
We find that generative tasks drift often due to the subjectivity of the answer space (76-89\%), compared to high-complexity tasks (7-21\%).
To identify the reasons, eight human experts analyze 170 multi-agent debates suffering from problem drift.
We find the most common issues related to this drift are the lack of progress (35\% of cases), low-quality feedback (26\% of cases), and a lack of clarity (25\% of cases).
We propose \judge, an LLM-as-a-judge method, as a first baseline to detect problem drift.
We also propose \policy, which mitigates 31\% of problem drift cases.
Our study is a step toward understanding a key limitation of multi-agent debate, highlighting why longer debates can harm task performance and how problem drift could be addressed.
\end{abstract}

\input{1_introduction}
\input{2_relatedWork}
\input{3_definition}
\input{4_methodlogy}

\input{5_experiments}
\input{6_epilogue}

\bibliography{25_ProblemDrift}

\appendix

\input{7_appendix}

\clearpage
\input{annotation.tex}

\end{document}

%% file: 1_introduction.tex
\section{Introduction}

Inspired by Social Choice Theory \citep{Endriss17}, recent research considers the use of multiple large language models (LLMs) to solve complex reasoning tasks and mitigate the limitations of single models, such as lacking modularity and answer diversity \citep{YinSCG23a, ChenSB24a, SunYLW24}.
Agents, embedded with varying expertise, memory, planning, and tool use \citep{ChenSB24a, SunYLW24, ZhuangYWS23a, BakerKMW20a}, can emulate human interaction to solve problems \citep{GuoCWC24a, Becker24}.
Recent work highlights the strengths of multi-agent debate (MAD) in reasoning and creativity \citep{ZhaoHXL23a, XuYLW23a, SuzgunK24a}.
MAD also scales test-time compute to solve challenging tasks similar to reasoning models, such as OpenAI o4 \citep{OpenAIo4} and DeepSeek R1 \citep{deepseekai2025deepseekr1incentivizingreasoningcapability}, which can be more effective than scaling training compute or data \citep{SnellLXK24a}.

\begin{figure}[t]
    \centering
    \includegraphics[width=0.99\linewidth]{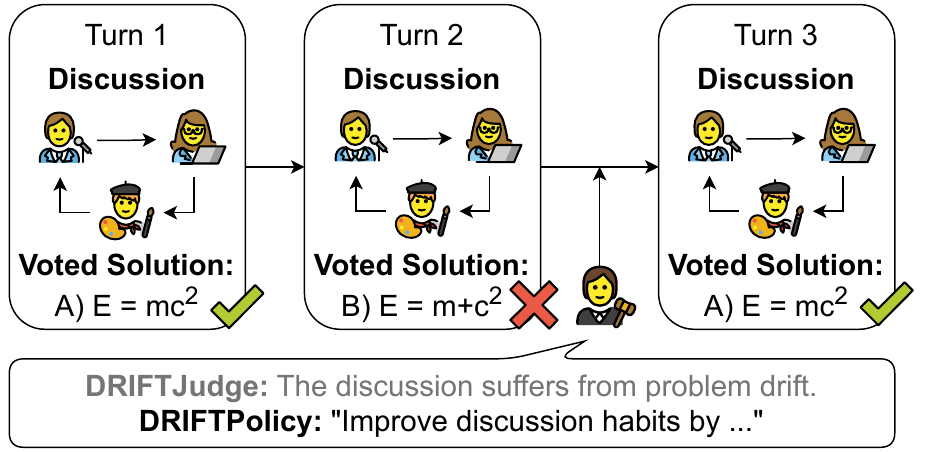}
    \caption{Problem drift in MAD. \judge\ detects problem drift at test-time. \policy\ provides on-demand feedback about the conversation.}
    \label{fig:problem_drift}
\end{figure}

However, the increasing complexity of MAD can promote errors and undesired behaviors (e.g., propagating hallucinations) \citep{GuoCWC24a}.
Problems originating from debate comprise agent orchestration \citep{WangWST24, ShahW24}, diminished planning capabilities \citep{ValmeekamMK23a}, and ineffective criticism generated by the LLM \citep{StechlyMK23a}.
Still, it is largely unknown what causes MAD to fail and why it happens.
It remains uncertain how MAD can avoid issues like ineffective criticism, which is crucial for high-quality debate and reliable results.
Few mitigation methods have been proposed for the few errors identified by related work.


This paper systematically analyzes errors that lead to performance degradation in MAD.
Through automated methods and a human evaluation, we observe that multi-agent systems can collapse in long discussions.
In some debates, agents' communication tends to deteriorate over time, drifting to a point where they can not recover and address the original task goal.
We refer to this phenomenon as \textbf{problem drift} and investigate its underlying causes and effects.
To identify problem drift post-hoc in MAD, we propose \metric, a simple yet effective metric that measures the quality of the ongoing discussion.
Our results suggest that problem drift occurs across all tested tasks, being most prevalent in generative tasks (76\%-89\% of samples), followed by instruction-following tasks (21\% of samples).
The majority of observed drifts do not recover.
Once a discussion drifts away, agents often do not reach the correct solution for a given task, except in 9\% of the translation and 45\% of the ethical question-answering examples. 
To identify and mitigate problem drift at test-time, we propose \judge\ and \policy.
While \judge\ identifies problem drift acting as an LLM-as-a-judge \citep{ZhengCSZ23}, \policy\ introduces a policy feedback agent \citep{FuPKL23b} that advises participating agents to improve the debate.
\Cref{fig:problem_drift} shows the dynamics between \judge\ and \policy\ during MAD.
We show that \policy\ can reduce the number of drifting discussions by 31\%, improving task accuracy by up to 3.6\% for weaker model agents.

By asking eight human experts, we identify eight reasons why problem drift occurs in MAD and categorize them into temporal error types (e.g., lack of progress) and local error types (e.g., task non-compliance).
Agent discussions are especially prone to a lack of progress (35\% of drifting samples) that can lead agents to overanalyze problems and give low-quality feedback.
Our work can be seen as a systematic analysis of MAD, showing the inherent limitations of prolonged agent interaction.
We release the code and data publicly\footnote{\mbox{\url{https://github.com/jonas-becker/problem-drift}}}.

%% file: 2_relatedWork.tex
\section{Related Work}

Ever since the first chatbots\footnote{The first recorded conversation between the chatbots ELIZA and PARRY is available here: \url{https://www.rfc-editor.org/rfc/rfc439}}, humans have been fascinated by the ability of computers to communicate in a human-like manner.
Recent advances in LLMs to reason and solve complex tasks \citep{OpenAIAAA24a} lead to a surge in studies on multi-agent systems \citep{ZhaoHXL23a, XuYLW23a, SuzgunK24a}.
\citet{GuoCWC24a} conduct a literature review on multi-agent LLMs.
Their taxonomy points to two main areas in which multi-agent LLMs are used: simulation and problem-solving.
Our investigation explores problem-solving because these tasks offer a controllable environment to probe components in multi-agent interaction \citep{YinSCG23a, DuLTT23}. 

\noindent{\textbf{Supportive Works.}}
Several works highlight the strengths of MAD, including divergent thinking \citep{degenofthought}, reasoning \citep{YinSCG23a}, creative writing \citep{SchickDJP22a}, dialogue generation \citep{ChenSB24a}, and theory of mind \citep{LiCSC23a}.
These works often use prompted personas \citep{WangMWG23a, XuYLW23a} and self-correction mechanisms like self-consistency \citep{WangWSL23} in a conversational setup.
However, the heterogeneous experimental setup across these studies (e.g., agent orchestration, decision-making, prompting) hinders their comparability \citep{Becker24, GuoCWC24a}.

\noindent{\textbf{Critical Works.}}
Other researchers study the limitations of multi-agent systems.
Work led by Microsoft and Salesforce recently showed that many LLMs perform worse in multi-turn settings on generative problems, suffering from a systematic performance degradation \citep{laban2025llmslost}.
\citet{WangWST24} and \citet{zhang2025multiagentdebateanswerquestion} focus on the computational cost of MAD, showing that single-agent LLMs can often achieve similar or even better performance through prompting.
Systems that include a self-critique mechanism (e.g., MAD) might diminish planning performance \citep{ValmeekamMK23a}. 
The correctness and content of LLM-generated criticism can be irrelevant to the performance of iterative prompting \citep{StechlyMK23a}.
\citet{ShahW24} argue that MAD has many issues, such as generalization, scalability, coordination, robustness, and ethical concerns.

\noindent{\textbf{Research Gap.}}
We highlight that MAD is relevant for problems that require divergent input or modular solutions.
While existing literature highlights diminished returns with prolonged MAD, the causes for such remain underexplored.
We address this gap by characterizing the deterioration of MAD performance (RQ1), its prevalence (RQ2), and whether agents can recover from it (RQ3).
We systematically study possible causes and effects through a human evaluation (RQ4) and propose a first baseline to detect (RQ5) and recover (RQ6) the degrading discussions.

%% file: 3_definition.tex
\section{Problem Drift} \label{sec:definition}

We introduce the concept of \textbf{problem drift} in MAD, which we define as the systematic decay in turn-based task performance as agents gradually diverge from the initial task. 
To quantify this effect, we propose \metric\ to measure the strength of the decay over a number of consecutive turns.

\paragraph{\textbf{\metric.}}
We describe how the quality of a solution differs in one turn from the previous turn.
Let $(x, y)$ be a pair of input and gold label for a task example in a dataset to be solved. 
A multi-agent system discusses for $N$ rounds.
At the end of each round $r \in (1, ..., N)$, the agents try to solve $x$ by providing an answer $\hat{y}^{(r)}$. 
Let $M(\hat{y}^{(r)}, y) \in [0,1]$ be a downstream performance metric that compares the ground-truth solution to the solution produced by the MAD, where $1$ corresponds to the solution and $0$ corresponds to a wrong solution.
Values between $0$ and $1$ can indicate partial solutions for non-binary solutions.
We define \textit{problem focus}, further called \metric, as:
\vspace*{2mm}
\begin{equation} \label{def:focus}
\small
FOCUS_r = M(\hat{y}^{(r)}, y) - M(\hat{y}^{(r-1)}, y) \in [-1,1]
\end{equation}

\noindent
If $FOCUS_r < 0$, the solution's quality degrades during round $r$. If $FOCUS_r > 0$, the solution improves.
\metric\ purposely depends on task-performance as an outcome-based score.
We also test a semantic-based alternative in \Cref{app:alternative_metric} but find two main limitations that make it unattractive for this work.

\paragraph{\textbf{Problem Drift.}}
We define that an ongoing discussion has \textit{problem drift} if the current solution obtains lower performance than the solution of a previous debate round over multiple rounds.
Let $M$ be the number of consecutive discussion rounds for one task example.
A debate has problem drift from round $1$ to $M$ of strength $\theta \in [-1,0]$ if:
\begin{equation} \label{def:drift}
\small
FOCUS_{1,M} = \sum_{r=1}^{M} FOCUS_r = \theta
\end{equation}
We define problem drift by the simple sum over \metric\ values to model the shift in task performance across debate rounds.

\paragraph{\textbf{Recovery.}}
We say an example recovered from problem drift when the discussion gets back to or improves upon the performance before the problem drift occurred.
We define a discussion as recovered from problem drift if $\exists N,M \text{ where } N > M \text{ s.t.}$:



\vspace*{-4mm}
\begin{equation} \label{def:recovery}
    \small
    \sum_{r=1}^{N} FOCUS_r >= 0 \quad \wedge \quad \sum_{r=1}^{M} FOCUS_r < 0
\end{equation}

%% file: 4_methodlogy.tex
\section{Methodology}

We explain the multi-agent environment for this study, our proposed detection and mitigation methods for problem drift, and the datasets and metrics.

\begin{figure*}[ht]
    \centering
    \includegraphics[width=0.99\linewidth]{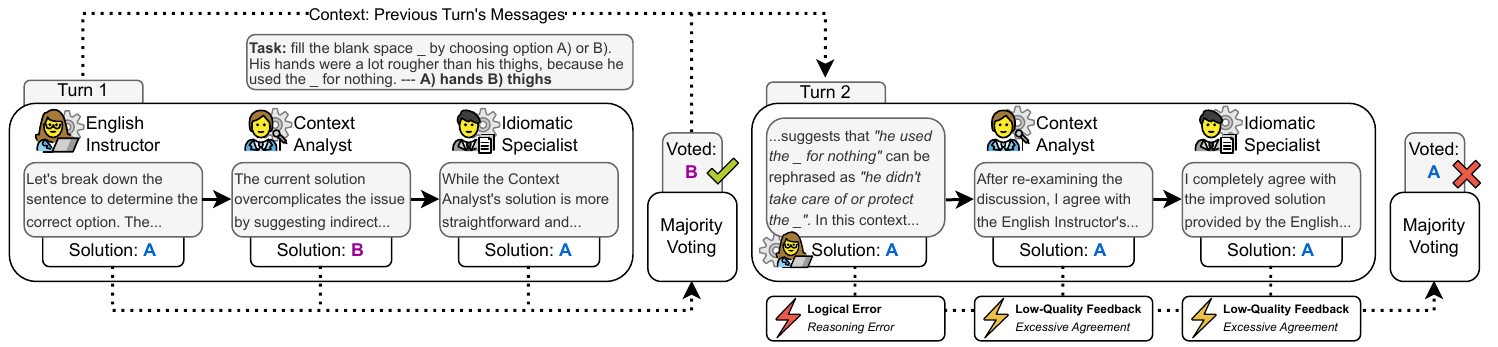}
    \caption{Example of problem drift in MAD. The \textit{English instructor} induces a logical error in the discussion. The other agents agree without skepticism, leading to the wrong solution and problem drift.}
    \label{fig:discussion_turn}
\end{figure*}

\subsection{Debate Setup} \label{sec:setup}
We define three components for running MAD: an interface for agents, discussion paradigms, and the decision-making protocol.
For each task, we create three agents with distinct \textbf{expert personas} \citep{XuYLW23a, ShiZWW23a}. 
We choose three agents as a hyperparameter following previous works \citep{ChenSB24a, YinSCG23a}.
These personas are automatically generated by \href{https://huggingface.co/meta-llama/Llama-3.1-70B-Instruct}{\textit{Meta-llama/Meta-Llama-3.1-70B-Instruct}} \citep{GrattafioriDJP24a} and induce a set of preferences (e.g., law attorney), which varies according to the task and sample (examples and details are in \Cref{app:examples}).
Second, we run a total of \textbf{seven turns} of conversation between the agents.
We find that in 99\% of cases, agents reach a first agreement within the first two turns.
Thus, we choose seven turns to capture the effects of prolonged agent interaction sufficiently.
This aligns with other setups for MAD \citep{YinSCG23a} and enables our analysis of recovery following problem drift.
Each agent can generate one new message at each turn and indicate their agreement with the current solution.
All agents see the messages of one another.
If an agent disagrees, it proposes a new solution concomitantly.
Finally, we employ two decision-making protocols that other researchers regularly use to ensure that our findings generalize across conversational setups \citep{YinSCG23a, kaesberg2025, YangDKH24}.
\textbf{Voting} is conducted between possible discussed solutions to reach a final solution for that turn \citep{YangDKH24}.
Following prior work, agents have access to one previous turn and the previously voted solution \citep{PagnoniBT21a, ZhangZ20b}.
A visual overview of the process is presented in \Cref{fig:discussion_turn}.
\textbf{Consensus} involves each agent directly modifying the current draft at their turn \citep{YinSCG23a}.
A visual overview of the process can be found in \Cref{fig:discussion_turn_consensus} of \Cref{app:overview_of_consensus}.


We conduct our experiments using \href{https://huggingface.co/meta-llama/Llama-3.1-70B-Instruct}{\textit{Meta-llama/Meta-Llama-3.1-70B-Instruct}} on eight NVIDIA A100 GPUs, and \href{https://huggingface.co/Qwen/Qwen2-7B-Instruct}{\textit{Qwen/Qwen2-7B-Instruct}} on two NVIDIA A100 GPUs with 40 GB each.
Detailed information about the framework \citep{becker2025mallmmultiagentlargelanguage}, parameters, and prompts is available in \Cref{app:framework,app:parameters,app:prompts}.
We report our main results using \textit{Llama-3.1} and \textit{Voting} but also include results for \textit{Qwen2} and \textit{Consensus} to ensure the generalizability of our findings.
Our goal is to show that problem drift and \metric\ are not exclusive to the specific design of a multi-agent system and apply to any MAD with intermediate solutions.
As the framework\footnote{\url{https://github.com/Multi-Agent-LLMs/mallm}} and the code\footnote{\url{https://github.com/jonas-becker/problem-drift}} are open source, the exact settings can be investigated, and all experiments can be reproduced.


\subsection{Mitigation Setup}
A natural question from our newly proposed definition of problem drift is whether we can mitigate it even when unsure about its presence (at test-time).

\subsubsection{Detection with \judge}
We aim to identify the occurrence of problem drift at test-time when gold labels are unknown.
We propose a first baseline to detect problem drift inspired by LLM-as-a-judge \citep{ZhengCSZ23}, which receives a focal turn's and a consecutive turn's solution to assess whether problem drift occurs.
To ensure that our results come from architectural changes instead of model capabilities, we refrain from using a powerful fine-tuned model as our judge.
We instead use the same model as for the agents with \judge.
This mitigation concept is independent of our model and could also use lightweight classifiers or models.

\subsubsection{Mitigation Methods}
We propose two mitigation methods to recover debates in the event of drift detection. 
We either \textbf{regenerate} the focal drifting turn or use a \textbf{\policy} agent that provides situational feedback.

\paragraph{\textbf{Regenerate.}} We undo the turn suffering from problem drift and regenerate that turn with a non-zero temperature. 
We only regenerate a turn once.

\paragraph{\textbf{\policy.}} We introduce a fourth agent at the end of the drifting turn that provides feedback about how to improve the current conversation and solve the issues leading to problem drift.
\policy\ is inspired by policy feedback agent by \citet{FuPKL23b} and the feedback mechanism used with error types in \citet{kirstein-etal-2025-whats}.
We include the prompt for our feedback generation in \Cref{app:prompts}.

\subsection{Datasets \& Metrics} \label{sec:datasetmetrics}

\paragraph{\textbf{Datasets.}} We select ten datasets from four domains: three generative tasks (XSum \citep{NarayanCL18b}, ETPC \citep{KovatchevMS18b}, WMT19 \citep{WikimediaFoundation19}), three reasoning tasks (StrategyQA \citep{GevaKSK21}, WinoGrande \citep{SakaguchiBBC19}, AQUA-RAT \citep{LingYDB17a}), three knowledge tasks (ETHICS \citep{HendrycksBBC23}, MMLU-Pro \citep{WangMZN24}, GPQA \citep{ReinHSP23a}), and one instruction following task (IFEval \citep{ZhouLMB23}).
With qualitative assessment, we identify StrategyQA, AQUA-RAT, and MMLU-Pro as \textit{complex} tasks, as they require a higher number of reasoning steps to be solved. 
The generative tasks ETPC, XSum, and WMT19 are characterized by \textit{subjectivity}, due to ambiguities or multiple valid solutions.
As MAD uses large amounts of test-time compute, it has become a common practice to evaluate subsets of datasets to study multi-agent systems \citep{YinSCG23a, ChenSB24a}.
We follow this approach.
Detailed information about the sampling process can be found in \Cref{app:datasets}.

\paragraph{\textbf{Metrics.}} We evaluate multiple-choice datasets (i.e., ETHICS, GPQA, MMLU-Pro, StrategyQA, WinoGrande, and Aqua-Rat) by accuracy.
For generative tasks (i.e., ETPC, XSum, and WMT19), we use BERTScore \citep{ZhangKWW19}.
We evaluate IFEval by the ``strict'' accuracy \citep{ZhouLMB23}.



\input{tables/good_bad_focus_discussions_heatmap}

%% file: tables/good_bad_focus_discussions_heatmap.tex
\begin{table*}[t]
\centering
\small
\begin{tabular}{cl|rr|rr|r}
\toprule
& & \multicolumn{1}{c}{\tiny{\# of samples staying at good perf.}} & \multicolumn{1}{c|}{\tiny{\# of samples staying at bad perf.}} & \multicolumn{1}{c}{\tiny{\# of improving samples}} & \multicolumn{1}{c|}{\tiny{\# of worsening samples}} & \multirow{2}{*}{\makecell[r]{\textbf{Total} \\ \textbf{Samples}}} \\
& \textbf{Dataset} & \textbf{$P(\hat{y}^{(r)}, y) > 0.7 \, \forall r$} & \textbf{$P(\hat{y}^{(r)}, y) < 0.7 \, \forall r$} & \textbf{$FOCUS_{1,7}>0$} & \textbf{$FOCUS_{1,7}<0$} & \\
\midrule
\parbox[t]{2mm}{\multirow{3}{*}{\rotatebox[origin=c]{90}{\tiny{Generative}}}} & ETPC & \cellcolor{green!2} 2.0\% \tiny{(22)} & \cellcolor{red!2} 1.9\% \tiny{(21)} & \cellcolor{green!26} 26.5\% \tiny{(287)} & \cellcolor{red!69} 69.5\% \tiny{(753)} & 1.083 \\
& XSum & \cellcolor{green!0} 0.3\% \tiny{(3)} & \cellcolor{red!8} 7.9\% \tiny{(91)} & \cellcolor{green!37} 37.9\% \tiny{(438)} & \cellcolor{red!54} 54.1\% \tiny{(626)} & 1.158 \\
& WMT19 & \cellcolor{green!14} 14.0\% \tiny{(143)} & \cellcolor{red!3} 3.0\% \tiny{(31)} & \cellcolor{green!13} 13.2\% \tiny{(135)} & \cellcolor{red!69} 69.9\% \tiny{(714)} & 1.023 \\
\midrule
\parbox[t]{2mm}{\multirow{3}{*}{\rotatebox[origin=c]{90}{\tiny{Reasoning}}}} & StrategyQA & \cellcolor{green!72} 72.2\% \tiny{(715)} & \cellcolor{red!18} 18.3\% \tiny{(181)} & \cellcolor{green!3} 3.7\% \tiny{(37)} & \cellcolor{red!5} 5.8\% \tiny{(57)} & 990 \\
& WinoGrande & \cellcolor{green!63} 63.4\% \tiny{(561)} & \cellcolor{red!21} 21.3\% \tiny{(188)} & \cellcolor{green!6} 6.8\% \tiny{(60)} & \cellcolor{red!8} 8.6\% \tiny{(76)} & 885 \\
& AQUA-RAT & \cellcolor{green!70} 70.5\% \tiny{(537)} & \cellcolor{red!21} 21.0\% \tiny{(160)} & \cellcolor{green!3} 3.8\% \tiny{(29)} & \cellcolor{red!4} 4.7\% \tiny{(36)} & 762 \\
\midrule
\parbox[t]{2mm}{\multirow{3}{*}{\rotatebox[origin=c]{90}{\tiny{Knowledge}}}} & GPQA & \cellcolor{green!33} 33.2\% \tiny{(225)} & \cellcolor{red!49} 49.9\% \tiny{(338)} & \cellcolor{green!8} 8.1\% \tiny{(55)} & \cellcolor{red!8} 8.7\% \tiny{(59)} & 677 \\
& MMLU-Pro & \cellcolor{green!51} 51.7\% \tiny{(578)} & \cellcolor{red!35} 35.7\% \tiny{(399)} & \cellcolor{green!4} 4.0\% \tiny{(45)} & \cellcolor{red!8} 8.7\% \tiny{(97)} & 1.119 \\
& ETHICS & \cellcolor{green!64} 64.0\% \tiny{(675)} & \cellcolor{red!22} 22.0\% \tiny{(232)} & \cellcolor{green!5} 5.7\% \tiny{(60)} & \cellcolor{red!8} 8.2\% \tiny{(86)} & 1.053 \\
\midrule
\rotatebox[origin=c]{90}{\tiny{IF}} & IFEval & \cellcolor{green!60} 60.4\% \tiny{(898)} & \cellcolor{red!21} 21.7\% \tiny{(323)} & \cellcolor{green!6} 6.1\% \tiny{(90)} & \cellcolor{red!11} 11.8\% \tiny{(175)} & 1.486 \\
\bottomrule
\end{tabular}
\caption{Percentage {\scriptsize (number)} of samples measured across debate rounds $r \in [1,\dots,7]$. Left cells: \hlgreen{green} indicates examples where \textit{performance remains $>0.7$} (second column); \hlred{red} indicates examples where \textit{performance remains $<0.7$} (third column). Right cells: \hlgreen{green} indicates examples improving over round 1 (\textit{positive \metric}, fourth column); \hlred{red} indicates examples degrading over round 1, i.e, problem drift (\textit{negative \metric}, fifth column). Cell opacity increases with percentage. MAD uses \textit{Llama-3.1} and Voting. \vspace{-0.5cm}}
\label{tab:good_bad_focus}
\end{table*}

%% file: 5_experiments.tex
\section{Experiments}

We present the experiments by answering a series of research questions about problem drift.

\noindent \textbf{RQ1}: \textit{How does the length of MAD impact task performance?} \textbf{Answer}: \textit{While some debates benefit from longer interactions, a notable subset degrades in performance, which we identify as problem drift.}


\Cref{tab:good_bad_focus} shows the performance trend of ongoing MAD.
We observe that some discussions (i.e., 3.7\%-37.9\%) benefit from longer MAD and improve performance compared to the first turn.
However, a significant chunk of discussions (i.e., 4.7\%-69.9\%) suffers from negative problem focus, leading to a performance drop during long debate.
While generative tasks (e.g., WMT19) show this in large quantities (54-69.9\%) as they are characterized by a subjective answer space, reasoning and knowledge tasks (e.g., StrategyQA, MMLU-Pro) display sporadic loss of focus (4.7-8.7\%).
However, a loss of focus on reasoning and knowledge tasks is more severe, as agents switch from a correct solution to a completely wrong solution.
Problem drift also concerns ethical alignment (8.2\%, ETHICS) and instruction-following (11.8\%, IFEval).
Notably, negative problem focus, i.e., problem drift, concerns all tasks to some extent.

Longer discussions provide more opportunities for intermediate errors.
This is relevant for two reasons.
First, these errors harm task performance in problem-solving and make benefits less reliable, questioning their cost-benefit ratio because of the computing requirements.
Second, if agents are unreliable during continuous debate, this raises concerns about unaligned behavior and harmful decision-making when employing autonomous agents in high-stake, human-centric applications \citep{HuaYJC24, MukobiRRS23}.
One might limit debates to a single turn to avoid problem drift altogether.
However, this would miss out on potential performance gain on 4-38\% of debates where $FOCUS_{1,7}>0$.
In addition, avoiding ongoing debate might be infeasible for some scenarios that are inherently multi-agent in nature and involve advanced reasoning chains or (semi-)autonomous multi-agent systems \citep{Slonim2021}.

In many cases, the performance during MAD remains stable, meaning it neither increases nor decreases over multiple rounds.
This is surprising, given the highlighted performance benefits of MAD in related work \citep{YinSCG23a, SchickDJP22a}.
As our later experiments will demonstrate, part of the success of MAD depends on individual discussion success and the fact that few discussions drift away from the problem.
Success also depends on the task and the conversational setup, such as agent orchestration and decision-making \citep{GuoCWC24a}, which may have been set carefully in related works.
We later show that problem drift universally occurs with consensus and voting decision-making and concerns different base models and model sizes (cf. \Cref{tab:task_performance}).

\noindent \textbf{RQ2}: \textit{How prevalent is problem drift, and does it depend on the task?} \textbf{Answer}: \textit{Generative tasks drift often due to the subjectivity of the answer space (76-89\%). 
Objective and high-complexity tasks drift less often (7-21\%) but more severely.}

\Cref{tab:turning_points} shows the statistics for samples suffering from problem drift on all datasets.
The column "Drifting Samples" shows the percentage of samples that have problem drift at any point during the debate.
The column ``Avg. Turns'' shows the average number of turns in a debate during which a solution drifts.
Multi-agent discussions for generative tasks (i.e., ETPC, XSum, WMT19) drift often, with XSum drifting for 74.6\% of samples and ETPC even drifting for 88.6\%.
We often observe agents proposing minor changes to the phrasing and wording of a solution.
Notably, generative tasks are discussed with the most turns where $FOCUS_r < 0$, i.e., when they drift.
Generative tasks are characterized by small incremental changes during MAD, leading to a small yet cumulative loss in \metric.
Knowledge and reasoning tasks drift between 6.6\%-15.5\% of samples.
The answers provided in a multiple-choice setup help the agents stay on track and drift less.
While task subjectivity impacts the occurrence of problem drift, it is not the only factor.

The few occurrences of problem drift on tasks like StrategyQA (6.6.\%) and AQUA-RAT (8.4\%) suggest that problem drift is not impacted by task complexity.
Possibly complex tasks that require several reasoning steps do not leave much room for the agents to debate unnecessary changes (e.g., aesthetics, rewording), leading to more valuable contributions by the agents.
Meanwhile, tasks characterized by subjectivity (e.g., ambiguities in WMT19) and less complex tasks (e.g., ETHICS) may suffer from agents overly contributing or providing meaningless information.
Our multi-agent setup was particularly prone to problem drift on the IFEval dataset (20.8\%).
Thus, agents are especially indecisive about how to adhere to instructions properly.
Agents often mention points unrelated to the intended tasks, causing problem drift by shallow or unhelpful feedback during the debate.

\input{tables/turning_points}

\noindent \textbf{RQ3}: \textit{Can discussions recover from problem drift?} \textbf{Answer}: \textit{Problems with lower task complexity have a high recovery rate of up to 45\%. Tasks that require complex reasoning and have a subjective answer space are recovered less with as low as 9\%.}

\input{tables/reasons}

The column "Recovery Rate" of \Cref{tab:turning_points} shows the percentage of samples that recover from problem drift during the debate.
The capability to recover concerns how many of the drifted samples perform on par or better than the performance before problem drift occurred (as detailed in \Cref{sec:definition}).
Debates on instruction-following (IFEval, 44.7\%) or ethical question-answering (ETHICS, 45.4\%) recover the most from problem drift.
The recovery rate is lower for more complex tasks (18.5\%-39.1\%) and subjective tasks (8.5\%-19.8\%).
For StrategyQA, which requires complex strategic planning to give the correct answers, only 18.5\% of drifting samples recover from problem drift.
For WMT19, a subjective translation task with ambiguities and multiple possible translations, only 8.5\% are recovered.
Problem drift is connected to both task complexity and subjectivity.
We also observe several cases where agents show redundant and circular behavior, overanalyze problems, and overly agree with wrong proposals.


\noindent \textbf{RQ4}: \textit{What are the possible reasons for problem drift as judged by human experts?} \textbf{Answer}: \textit{Both local and temporal errors appear with problem drift. Most often, a lack of progress leads to agents providing low-quality feedback and lacking clarity.}

We sample 170 examples that suffer from problem drift using \metric\ and follow a two-step approach to identify the reasons for problem drift with eight human experts.
First, we create a set of error types by automatically generating them from a large number of drifting conversations using \href{https://huggingface.co/meta-llama/Llama-3.1-70B-Instruct}{\textit{meta-llama/Meta-Llama-3.1-70B-Instruct}} \citep{GrattafioriDJP24a} and summarizing with ChatGPT-4o\footnote{Version: 3rd of December 2024.} and manual assessment.
Second, eight human experts annotate a subset of the 170 drifting debates to identify the relevance of each error type.
We leave space for additional explanation of the labels.
For the systematic and human assessments, we extract the relevant part of the discussion to keep context length manageable, i.e., the three messages of the turn before the drifting turn and the three messages of the driving turn.
This is reasonable because, during the debate, the agent's context memory is also limited to one turn.
Details on the annotation procedure are available in \Cref{app:dataset_annotation}.
We publicly release \dataset, a dataset with 170 samples, together with the human annotations and annotation guidelines, to our repository.

We classify the error types into \textit{local errors} (self-contained within a single message) and \textit{temporal errors} depending on contextual relationships between messages.
\Cref{tab:reasons} shows the error types that appear in conjunction with problem drift as well as the number of samples rated by human experts into that error type category.
The most common error type annotated is a lack of progress in 60 of the 170 samples.
During long conversations, agents often repeat each other's arguments, missing out on bringing new ideas to the discussions.
We find this relevant to Degeneration-of-Thought, as proposed by \citet{degenofthought}, which describes how single LLMs fail to generate novel thoughts during continuous self-reflection.
However, a lack of progress does not directly cause problem drift because, for a decrease in performance, changes to the final answer are needed.
Upon further investigation, we find that a lack of progress often occurs with other error types (e.g., low-quality feedback with 45 cases and lack of clarity with 43 cases).
When agents converge on a solution, and the conversation becomes redundant, they tend to propose small incremental changes to the solution that harm task performance.
Our human evaluation suggests that overanalysis and hypercritical suggestions are highly related to problem drift (e.g., agents struggle to choose the appropriate scope of their criticism).
We also notice cases where agents do not comply with task requirements (e.g., results contain multiple possible answers).
This is very present on the IFEval dataset, where instruction-following is the primary goal.
One possible reason why task compliance suffers during the debate is that personas amplify the agents' preferences, valuing individual preferences above task compliance in some cases.

The MAD also produces 23 linguistic errors within the discussion excerpts, such as typos or grammatical errors.
We find that sometimes agents get stuck in a repetitive generation.
This finding is surprising, as today's LLMs are considered to be highly competent regarding linguistic capabilities \citep{GrattafioriDJP24a}; this raises questions about how improvements in reasoning can come at the cost of degradation of linguistic quality, as also seen in recent work from DeepSeek-R1-Zero \citep{deepseekai2025deepseekr1incentivizingreasoningcapability}.
As we use the default parameters for our model (cf. \Cref{app:parameters}), we expect the causality for this to be unrelated to our specific model setup.
Potentially, prompting style or discourse relationships between messages could be confounding factors.

Human experts only selected ``Other'' as a label eight times, indicating that our provided list is conclusive.
In 29/170 cases, annotators chose ``None of these'', which we interpret as there being no clear error of a provided category in the snippet, not as evidence that the underlying task score did not decrease.
This makes the 17\% figure an upper bound on disagreement between human judgements and our stricter, metric-based notion of drift.
It does not correspond to a 17\% false-positive rate of FOCUS over the whole corpus, nor does it affect our main quantitative statements, which depend only on task metrics.
We include examples of all error types in \Cref{app:examples}.

\noindent \textbf{RQ5}: \textit{How well can identify problem drift at test-time when gold labels are unknown?} \textbf{Answer}: \textit{Our \judge\ has high specificity (0.92) but moderate recall (0.57) and low precision (0.15).}

We report the performance of \judge\ on the MMLU-Pro dataset, indicated by specificity (0.92), precision (0.15), recall (0.57), and accuracy (0.91).
Even though \judge\ achieves high accuracy, we see room for improvement in detecting problem drift with high precision.
Minimizing false positives with \judge\ is critical because \judge\ should not trigger any mitigation if the discussion is not actually drifting.
This paper provides an initial baseline for others to detect problem drift.
Fine-tuning a model on local and temporal error types (cf. \Cref{tab:reasons}) may help build a more robust detector.

\noindent \textbf{RQ6}: \textit{Can we mitigate problem drift when we know that it occurs?} \textbf{Answer}: \textit{Our \policy\ can effectively mitigate 30.6\% of drifting examples, improving task performance of prolonged debate.}

\input{tables/successful_samples_abs}

\Cref{tab:successful_samples} shows the number of samples that "Drift", samples that "Never Drift", and the percentage "Ratio" of samples that never drift compared to the total number of samples.
To quantify the effectiveness of our mitigation independent from detection performance, we use \metric\ to determine when problem drift occurs and invoke the mitigation method.
Using the \textit{Regenerate} mitigation, the ratio of never drifting discussions is 90.4\%, compared to 86.9\% using our multi-agent baseline without any mitigation.
With \policy\, which provides targeted feedback given the identified issues leading to problem drift of RQ4 (cf. \Cref{tab:reasons}), the ratio improves to 89.4\%.
The capability of \policy\ comes from prompting the feedback agent with the identified error types (cf. \Cref{tab:reasons}).

Task performance indicates that either mitigation method (\policy\ and \textit{Regenerate}) combined with \judge\ can reduce the performance drop in ongoing MAD.
Debating for seven turns leads to an average performance drop of \mbox{-4.65\%} accuracy for voting decision-making.
Voting decisions benefits most from our mitigation strategies.
Using \judge\ and \policy\ with weaker model agents (i.e., \textit{Qwen-2-7b}), we mitigate the performance drop from -3.66\% to -0.08\% accuracy.
Interestingly, improving performance through \policy\ is more difficult if models are already very capable.
Here, regenerating the drifting turn is the better solution, improving the delta from -4.65\% to -2.59\% accuracy.
The performance improvement appears incremental, which suggests a cost-benefit analysis.
We point out that employing \policy\ is more cost-efficient than regenerating the conversation round, only requiring one occasional extra message to be generated.
Compared with default MAD, using \judge\ and \policy\ combined yields a marginal 8.5\% increase in computational cost.
The 8.5\% is measured on top of an already expensive seven-turn debate, highlighting that, if computational cost is a concern, solutions outside the MAD paradigm would typically be considered.
More details on performance are provided in \Cref{app:task_performance}, and the computational analysis in \Cref{app:compute}.

The goal is to improve the mitigation of problem drift further, so that MAD can maintain high \metric\ to increase task performance.
Until then, incremental improvements are still relevant because limiting MAD to one turn might be infeasible, e.g., for (semi-)autonomous systems \citep{Slonim2021} or due to missing out on performance gains on 4-38\% of examples (cf. RQ1).
The scope of this work is to understand where problem drift occurs, what the reasons are, and if recovering from it is realistic.
Although both \judge\ and \textit{Regenerate} partially address problem drift, this is not a solved issue, as not all problem drift cases are mitigated.
One interesting direction is to fine-tune agents to make high-quality contributions to MAD, which could also involve our proposed metric \metric.
Negative-aware fine-tuning has shown recent success for reasoning tasks \citep{wang2024learningfailureintegratingnegative}, where our proposed dataset \dataset\ with human-annotated error types could be useful.

%% file: tables/turning_points.tex
\begin{table}[t]
\centering
\small
\begin{tabular}{cp{1.6cm}|rrr}
\toprule
& \textbf{Dataset} & \textbf{\shortstack{\tiny{Drifting} \\ \tiny{Samples (\%)}}} & \textbf{\shortstack{\tiny{Recovery} \\ \tiny{Rate (\%)}}} & \textbf{\shortstack{\tiny{Avg. Turns} \\ \tiny{\scalebox{.7}{$FOCUS_r < 0$}}}} \\
\midrule
\parbox[t]{2mm}{\multirow{3}{*}{\rotatebox[origin=c]{90}{\tiny{Generative}}}} & ETPC & \textbf{88.6 }\tiny{$\pm 0.8$} & 19.8 \tiny{$\pm 1.8$} & \textbf{2.23\tiny{$\pm 0.04$}}\\
& XSum & 74.6 \tiny{$\pm 0.7$} & 19.0 \tiny{$\pm 1.0$} & 1.29\tiny{$\pm 0.07$}\\ 
& WMT19 & 76.3 \tiny{$\pm 2.7$} & \textbf{8.5 }\tiny{$\pm 0.3$} & 1.42\tiny{$\pm 0.05$}\\
\midrule
\parbox[t]{2mm}{\multirow{3}{*}{\rotatebox[origin=c]{90}{\tiny{Reasoning}}}} & StrategyQA & \textbf{6.6 }\tiny{$\pm 0.6$} & 18.5 \tiny{$\pm 5.7$} & \textbf{0.07\tiny{$\pm 0.00$}}\\ 
& WinoGrande & 14.9 \tiny{$\pm 1.8$} & 39.4 \tiny{$\pm 4.8$} & 0.17\tiny{$\pm 0.03$}\\
& AQUA-RAT & 8.4 \tiny{$\pm 2.1$} & 39.1 \tiny{$\pm 4.4$} & 0.09\tiny{$\pm 0.02$}\\
\midrule
\parbox[t]{2mm}{\multirow{3}{*}{\rotatebox[origin=c]{90}{\tiny{Knowledge}}}} & GPQA & 13.3 \tiny{$\pm 1.1$} & 23.3 \tiny{$\pm 3.0$} & 0.14\tiny{$\pm 0.01$}\\
& MMLU-Pro & 13.1 \tiny{$\pm 0.6$} & 25.9 \tiny{$\pm 7.8$} & 0.14\tiny{$\pm 0.00$}\\
& ETHICS & 15.5 \tiny{$\pm 1.2$} & \textbf{45.4 }\tiny{$\pm 5.4$} & 0.18\tiny{$\pm 0.01$}\\
\midrule
\rotatebox[origin=c]{90}{\tiny{IF}} & IFEval & 20.8 \tiny{$\pm 1.7$} & 44.7 \tiny{$\pm 4.9$} & 0.25\tiny{$\pm 0.02$}\\
\bottomrule
\end{tabular}
\caption{Statistics showing the percentage of \textbf{Drifting Samples} at any point in the debate, the percentage of \textbf{Recovering Samples} relative to the drifting ones, and the average number of turns after which problem drift occurs \textbf{$FOCUS_r \neq 0$}. The highest and lowest values per column are displayed in \textbf{bold}. Values are reported with their respective standard deviations for three runs. MAD uses \textit{Llama-3.1} and Voting.\vspace{-0.5cm}}
\label{tab:turning_points}
\end{table}

%% file: tables/reasons.tex
\begin{table*}[t]
\centering
\small
\begin{tabular}{p{2cm}rp{8cm}r}
\toprule
\textbf{Category} & \textbf{Error Type} & \textbf{Explanation} & \textbf{Cases}\\
\midrule
 & Lack of Progress & \tiny{Inefficiency, Redundancy, Circular discussion, Repetition, Unproductive disagreement} & 60 \\
Temporal & Low-Quality Feedback & \tiny{Excessive criticism, Excessive agreement, Self-contradictory feedback, Unhelpful feedback} & 45\\
 & Low-Quality Engagement & \tiny{Poor collaboration, Minimal participation, Disjointed contribution, Ignorance} & 25 \\
\midrule
 & Lack of Clarity & \tiny{Overanalysis, Overgeneralization, Insignificant changes} \hspace{5cm} & 43 \\
 & Task Non-Compliance &  \tiny{Off-topic, Bad instruction following} & 35\\
Local & Knowledge Gap & \tiny{Assumptions, Lack of data, Hallucinated facts, Wrongly cited} & 28 \\
 & Logical Error & \tiny{Lack of common sense, Reasoning error} & 28 \\
 & Linguistic Error & \tiny{Fluency, Grammatical errors, False pronouns} & 23 \\
\midrule
 & None of these & \tiny{No problem drift} & 29 \\
 & Other & \tiny{Any other error} & 8 \\
\bottomrule
\end{tabular}
\caption{Error types that occur with problem drift. Eight human experts assessed a total of 170 discussion excerpts to determine how frequently each error type occurs.\vspace{-0.5cm}}
\label{tab:reasons}
\end{table*}

%% file: tables/successful_samples_abs.tex
\begin{table}[t]
\centering
\small
\begin{tabular}{l|rr|r}
\toprule
\textbf{Mitigation} & \textbf{Drift}  & \textbf{Never Drift} & \textbf{Ratio} \\
\midrule
None (MAD) & 49.0\tiny{$\pm 2.2$} & 324.0\tiny{$\pm 2.2$} & 86.9\%\tiny{$\pm 0.6$} \\
DRIFTPolicy & 39.7\tiny{$\pm 6.2$} & 333.3\tiny{$\pm 6.2$} & 89.4\%\tiny{$\pm 1.7$} \\ 
Regenerate & \textbf{35.7\tiny{$\pm 2.1$}} & \textbf{337.3\tiny{$\pm 2.1$}} & \textbf{90.4}\%\tiny{$\pm 0.6$} \\
\bottomrule
\end{tabular}
\caption{Number of samples on MMLU-Pro listed by mitigation method that show or never show problem drift during MAD. The last column refers to the percentage of never drifting out of all samples. Better values are displayed in bold. MAD uses \textit{Llama-3.1} and Voting. \vspace{-0.75cm}}
\label{tab:successful_samples}
\end{table}

%% file: 6_epilogue.tex
\section{Conclusion}

In this work, we identified \textbf{problem drift}, a performance degradation in MAD.
We conducted experiments on ten datasets across ten different tasks to quantify problem drift and found that it occurs across all tasks.
Our experiments showed that problem drift affects 7\% (StrategyQA) to 89\% (ETPC) of debates, but especially when discussing low-complexity and generative problems.
We proposed \judge\ and \policy\ as first baselines to detect and mitigate problem drift.
\policy\ reduced occurrence of problem drift, mitigating 30.6\% of problem drift for weaker agents.
We characterized eight error types through a human evaluation with eight human participants.
These errors were grouped into two main categories: temporal (i.e., lack of progress, low-quality feedback, low-quality engagement) and local (i.e., lack of clarity, task non-compliance, knowledge gap, logical error, linguistic error).
We published the code, the \dataset\ dataset, and the human annotations.



Future work could explore the role of agents in discussing new ideas that are not directly task-related but help to reason deeply.
Other interesting directions lie in comparing the inner dynamics of human and agent debates, such as differences in explorativeness, conciseness, and argument structure.
One can question whether problem drift is a natural property that multi-agent systems present when exploring new ideas and solutions, similar to how humans explore different reasoning paths in discussions.
Our findings suggest that, in some cases, agents drift away to recover later on in the discussion. 
Still, these agents regularly become prone to simple temporal and local mistakes.

\section*{Limitations}

The large amount of generated text in MAD can hinder qualitative human analysis of their discussions, as annotation is expensive.
In this human study, we investigated only the drifting turn and its immediate predecessor.
This decision provides a reasonable space to search for local and short-context temporal mistakes agents make during the debate. 
Still, it does not consider the full discussion history, which may include valuable information.
To keep our findings on error types comparable to the actual MAD, we also limit the context length of the agents to the current and previous turns during all experiments.

Our \judge\ requires forwarding solution pairs from consecutive turns, which can be computationally expensive. 
The combination of \judge\ and \policy\ yields an 8.5\% increase in computational cost (cf. \Cref{tab:compute} of \Cref{app:compute}).
We did not find this limiting for this project, but note that for production environments, fine-tuning a more cost-efficient classifier like BERT \citep{sun2020finetuneberttextclassification} on drifting and non-drifting examples could be a promising alternative.


\section*{Acknowledgements}
This work was partially supported by the Lower Saxony Ministry of Science and Culture and the VW Foundation. 
We acknowledge EuroHPC Joint Undertaking for awarding us access to MeluXina at LuxProvide, Luxembourg.
This work was funded by the Deutsche Forschungsgemeinschaft (DFG, German Research Foundation) – 564661959.
Many thanks to Dominik Meier for the thoughtful discussions.

%% file: 7_appendix.tex
\section{MALLM Framework} \label{app:framework}

We provide more thorough details about the framework used to run the experiments called MALLM (\textbf{M}ulti-\textbf{A}gent \textbf{L}arge \textbf{L}anguage \textbf{M}odels) \citep{becker2025mallmmultiagentlargelanguage}.

\paragraph{\textbf{Automatic Persona Assignment.}}
Discussions with MALLM use task- and example-specific personas for the agents.
As manually specifying useful personas for each example is not feasible, we automatically assign personas that foster a rich discussion.
For this, we explicitly prompt another LLM (\textit{meta-llama/Meta-Llama-3.1-70B-Instruct}) to generate a diverse set of three expert personas for each example.
This yields a set of experts that represents various beliefs, opinions, and proficiency.
The prompt for the automatic persona assignment can be read in \Cref{app:prompts}.
My approach follows previous works like Solo-Performance-Prompting \citep{WangMWG23a} and Meta-Prompting \citep{SuzgunK24a}, which show that existing LLMs can be leveraged to generate and consult fitting personas on a problem automatically.
We use three agents for this study, following previous works \citep{ChenSB24a, YinSCG23a} because the structural complexity is better than with two agents while not being too complex to provide meaningful insights.
While other works have used different types of personas like personalities \citep{ShiZWW23a}, the personas generated in this work are experts related to the task and example.

\paragraph{\textbf{Discussion Paradigm.}}
To define the structure of the multi-agent discussion, we use the \textit{memory} paradigm defined by \citet{YinSCG23a}.
\Cref{fig:problem_drift} shows the format graphically.
With the memory paradigm, all agents contribute to the discussion once per turn and have all the information available.
While there are potentially many ways to define the discourse's structure, we choose this paradigm because of its simplicity, keeping the human assessment of agent discussions manageable.

\paragraph{\textbf{Voting.}}
A voting mechanism at the end of each turn allows for a fixed discussion length while still providing a solution to evaluate at each turn.
We select this iterative voting approach because it universally fits our diverse selection of generative and QA tasks.
The prompt for the voting is visible in \Cref{app:prompts}.
In the case that the agents produce a tie by voting, we select a random one.
This only happened with 0.65\% of all voting procedures.
Using voting, this study differs from other works that either employ no decision-making at all \citep{SchickDJP22a} or use a judge agent that makes the final decision \citep{SunYLW24}.
Meanwhile, this voting approach yields a definitive solution to evaluate after each turn.

\section{Parameters} \label{app:parameters}

We adhere to default parameters for our used models, using langchain 0.1.16 and openai 1.25.0 for the implementation of the MALLM framework.

\begin{itemize}[noitemsep]%
    \item \texttt{temperature = 1.0}
    \item \texttt{top\_p = 1.0}
    \item \texttt{presence\_penalty = 0.0}
    \item \texttt{frequency\_penalty = 0.0}
    \item \texttt{max\_tokens = 1024}
\end{itemize}

\section{Alternative FOCUS Metric}\label{app:alternative_metric}

Our proposed metric $FOCUS$ uses task performance as an indicator for drift (i.e., when longer discussions reduce in performance, that is an indicator of drift).
Initially, we had also considered a definition of $FOCUS$ that relies on the semantic similarity of discussions, called $FOCUS^{Sim}$.
$FOCUS^{Sim}$ uses the semantic similarity of contextual embeddings between turn messages and the task instruction.
First, we compute the contextual embedding of the task description with a distilled version of BERT \citep{devlin2019bertpretrainingdeepbidirectional} called \texttt{all-MiniLM-L6-v2}.
At each turn, we average the contextual embedding of all three contributed messages and compute the cosine similarity of the averaged embedding to the embedding of the task description.

\input{tables/compare_focus_variants}

\Cref{tab:compare_focus_variants} compares both variants of $FOCUS$.
We see that both metrics indicate the strongest drift at the beginning of the debate (-2.32 for $FOCUS$ and -6.86 for $FOCUS^{Sim}$).

We did not include this semantic variant primarily due to two main issues. First, semantic embeddings are often limited in their ability to capture complex reasoning. A discussion with some semantic variability does not necessarily mean that the discussion is leading to worse outcomes. Accuracy is outcome-based, which is more fitting in many contexts (e.g., agents can discuss various semantic parts of the question individually, but the final answer aggregating these views is correct, showing low outcome-based drift). Second, semantic drift is biased towards individual error types (e.g., task non-compliance) from our human study.
We further observe that both metrics show little to no correlation (Spearman $\rho=0.024$, $p<0.05$).
Because we also aim to identify the reasons for problem drift in this work, it can not replace our original metric $FOCUS$.
The investigation of a task-independent quantification of $FOCUS$ and problem drift remains a possible direction for future work.

An alternative approach to measuring problem drift could favor more recent rounds. 
However, this would penalize temporary explorations that can be later recovered. 
In contrast, a momentum term would not accurately capture the severity of problem drift at specific rounds.

\section{Prompts} \label{app:prompts}

\input{prompts/discussion}
\input{prompts/solutionExtraction}
\input{prompts/voting}
\input{prompts/personaGenerator}
\input{prompts/llmJudge}
\input{prompts/policyFeedback}

\onecolumn

\section{Datasets} \label{app:datasets}

\input{tables/datasets}

As discussions require many tokens to be generated and computing resources are limited, only subsets of the datasets are evaluated.
We sample a subset of size $n_{\text{subset}}$ from each dataset for our experiments by a 95\% confidence interval and a 5\% margin of error (MoE), conservatively assuming a sample proportion $p=0.5$ \citep{Cochran53}.
\begin{equation} \label{eq1}
\begin{split}
n &=  \frac{Z_{0.975}^2 \cdot p(1 - p)}{\text{MoE}^2} \\ 
n &=  \frac{1.96^2 \cdot 0.5(1 - 0.5)}{0.05^2} = 384.16 \approx 385 \\
n_{\text{subset}} &= \frac{n}{1 + \left(\frac{n - 1}{N_{\text{dataset}}}\right)} = \frac{385}{1 + \left(\frac{385 - 1}{N_{\text{dataset}}}\right)} \\
\end{split}
\end{equation}
This yields several hundred samples per dataset as our test sets with a 95\% confidence interval and 5\% margin of error.
Several other studies on multi-agent systems also evaluate a subset of discussions \citep{YinSCG23a, ChenSB24a}.
To further quantify if the results reflect the complete datasets, we follow \citet{WangPSB24a} and run each experiment three times on randomized subsets and report the standard deviation of our results between the runs.

\section{Examples} \label{app:examples}


\subsection{Lack of Progress}
Lack of progress comprises inefficiency, redundancy, circular discussion, repetition, and unproductive disagreement.
In the example, the agents repeat the reasons for an unnecessary change (changing a word from "statement" to "assertion") without bringing novel ideas to the debate.
\input{examples/tex/lack_of_progress}

\subsection{Low-Quality Feedback}
Low-quality feedback comprises excessive criticism, excessive agreement, self-contradictory feedback, and unhelpful feedback.
In the example, the pop culture enthusiast suggests a minor change to make the answer clearer (excessive criticism), but fails to follow the instruction as before.
\input{examples/tex/low_quality_feedback}

\subsection{Low-Quality Engagement}
Low-quality engagement comprises poor collaboration, minimal participation, disjointed contribution, and ignorance. 
In the example, the anthropologist does not touch upon any points previously made by the other agents. Instead, it introduces a separate perspective and solution.
\input{examples/tex/low_quality_engagement}

\subsection{Lack of Clarity}
Lack of clarity comprises overanalysis, overgeneralization, and insignificant changes.
In the example, the voted solution was correct initially, but the language expert overly analyzes the circumstances, as Mrs. Fairfax could have shared more information about Mr. Rochester.
\input{examples/tex/lack_of_clarity}

\subsection{Task Non-Compliance}
Task non-compliance comprises off-topic and bad instruction following.
In the example, the agents fail to stick to one of the provided options of the multiple-choice task. Instead, they invent a new variant as an answer.
\input{examples/tex/task_compliance}

\subsection{Knowledge Gap}
Knowledge gap comprises assumptions, lack of data, hallucinated facts, and wrongly cited facts.
In the example, the agents refer to the China GSHS 2015 report. However, the report for that year does not exist.
\input{examples/tex/knowledge_gap}

\subsection{Logical Error}
Logical errors comprise a lack of common sense and reasoning errors.
In the example, the reading comprehension specialist says, "If the fan were easier to access, it would be dusted more often," but fails to realize that this means that the dresser is easier to reach.
\input{examples/tex/logical_error}

\subsection{Linguistic Error}
Linguistic errors comprise fluency, grammatical errors, and false pronouns.
In the example, the university student and the statistician fail to produce fluent text and get stuck in a loop of text generation.
\input{examples/tex/linguistic_error}

\clearpage

\section{Task Performance} \label{app:task_performance}
\input{tables/task_performance}

\input{tables/task_performance_delta}

\input{tables/good_bad_focus_discussions_heatmap_gpt-5-mini}

\clearpage

\section{Overview of Consensus} \label{app:overview_of_consensus}
\begin{figure*}[h]
    \centering
    \includegraphics[width=0.99\linewidth]{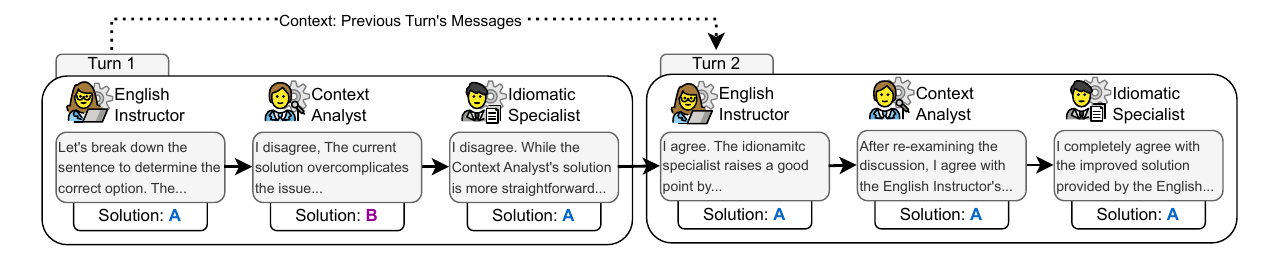}
    \caption{Example of consensus-based MAD. The draft of the solution is continuously improved without any voting in between turns.}
    \label{fig:discussion_turn_consensus}
\end{figure*}

\clearpage

\section{Dataset Annotation} \label{app:dataset_annotation}

\subsection{Annotation Instructions}
\begin{figure}[h]
    \centering
    \includegraphics[width=0.99\linewidth]{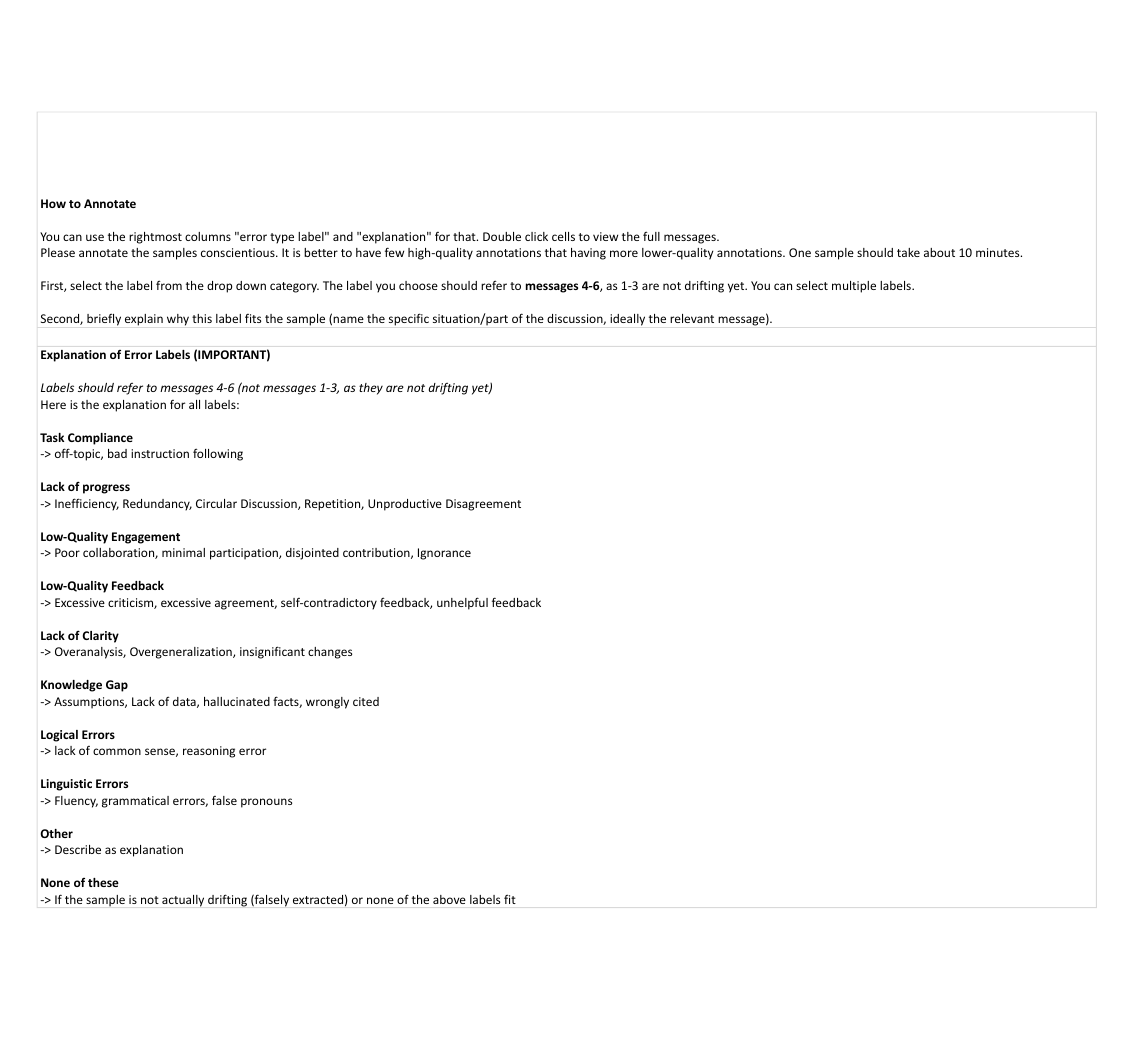}
    \caption{Instructions for the human annotation.}
    \label{fig:inst_human_annot}
\end{figure}

\clearpage

\twocolumn

\subsection{Annotation Procedure}

\paragraph{\textbf{Systematic Reason Extraction}}
Because of the scale and amount of multi-agent conversations, we automatically extract 200 examples (20 per dataset) that suffer from problem drift, i.e., where turn 7 performance is worse than turn 1, and ask \textit{meta-llama/Meta-Llama-3.1-70B-Instruct} to provide a reason for the problem drift.
The prompt for this can be found in \Cref{app:prompts}.
We then ask ChatGPT-4o\footnote{Version: 3rd of December 2024} to summarize all 200 reasons into a manageable list of 26 labels.
This method builds on previous studies that successfully utilize LLMs during the annotation process \citep{WangKRM24, KimMCR24}.

\paragraph{\textbf{Human Assessment}}
To obtain a concise set of factors that could cause problem drift, we manually check the labels generated during the first step for their orthogonality.
In this process, we eliminate redundant entries and sort other reasons into a total of nine resulting categories.
The full list of categories can be seen in \Cref{tab:reasons}.
Next, we ask eight human experts in natural language processing to jointly annotate a total of 170 drifting discussions using the created list of errors.
The annotators are three doctoral students and five research assistants.
One of them is female and the others are male, aged from 21 to 31.
We ask them to select one or multiple of the provided error types based on the messages and solutions of the two turns relevant to problem drift (i.e., the turn before the drift and the turn causing the drift).
We set this limit to keep the context manageable for humans.
Each expert annotates a share of the work.
Due to the effort and associated cost of the complex annotation procedure, there is no overlap between the annotator's samples.
However, annotators should provide a brief explanation of why the labels were selected for the sample.
We also include these explanations for the final dataset.
Assuming that our systematic extraction could have left an important error type out of our final set, we provide a separate field ``Other'' to the annotators with the opportunity to explain.
We also provide the field "None of the above" to capture samples that might have been falsely extracted as drifting samples (e.g., due to evaluation or dataset noise).
We publicly release the produced dataset with 170 samples called \textbf{DriftEval} together with the human annotations to our repository.

\section{Computation} \label{app:compute}

\input{tables/compute}

\section{Usage of AI}

In the conduct of this research project, we used specific artificial intelligence tools and algorithms (ChatGPT-4o and Grammarly) to assist with writing, coding, and aggregating experiment data. While these tools have augmented our capabilities and contributed to our findings, it's pertinent to note that they have inherent limitations. We have made every effort to use AI in a transparent and responsible manner. Any conclusions drawn are a result of combined human and machine insights. This is an automatic report generated with © AI Usage Cards \citep{wahle2023aiusagecard}.

%% file: tables/compare_focus_variants.tex
\begin{table}[t]
\centering
\small
\begin{tabular}{r|rr}
\toprule
\textbf{Turn} & \textbf{$FOCUS_{r}$} & \textbf{$FOCUS^{Sim}_{r}$} \\
\midrule
1 & - & - \\
2 & \textbf{-2.32\tiny{$\pm 1.3$}} & \textbf{-6.86\tiny{$\pm 0.4$}} \\
3 & -0.98\tiny{$\pm 0.9$} & -1.78\tiny{$\pm 0.2$} \\
4 & -0.54\tiny{$\pm 0.4$} & -0.64\tiny{$\pm 0.1$} \\
5 & -0.27\tiny{$\pm 0.6$} & -0.77\tiny{$\pm 0.1$} \\
6 & -0.45\tiny{$\pm 0.6$} & -0.45\tiny{$\pm 0.2$} \\
7 & -0.63\tiny{$\pm 0.3$} & -0.29\tiny{$\pm 0.2$} \\
\bottomrule
\end{tabular}
\caption{Comparison of $FOCUS$ and $FOCUS^{Sim}$ across seven turns for the MMLU-Pro dataset. The strongest values are highlighted. Values are reported with their respective standard deviations for three runs.}
\label{tab:compare_focus_variants}
\end{table}

%% file: prompts/discussion.tex
\begin{figure}[H]
\scalebox{0.5}{%
    \begin{tikzpicture}
    \node [draw, rectangle, rounded corners, fill=gray!20, text width=0.95\textwidth, inner sep=10pt] (block) {
        \begin{minipage}{\textwidth}
        \textbf{You take part in a discussion to solve a task.} \\

        \textbf{Task:} \texttt{<instruction>} \\
        \textbf{Input:} \texttt{<example>} \\
        \textbf{Context:} \texttt{<optional information>} \\
        \textbf{Your role:} \texttt{<persona name>} (\texttt{<persona description>}) \\
        \textbf{Current Solution:} \texttt{<most recent draft>}

        \textbf{This is the discussion to the current point:} \\
        \texttt{<agent memory>} \\

        \textbf{Improve the current solution. If you agree with the current solution, answer with [AGREE], else answer with [DISAGREE] and explain why and provide an improved solution.} \\
        \textbf{Let's think step-by-step.}
        \end{minipage}
    };
    \end{tikzpicture}
}
\caption{Prompt to an agent that contributes to the discussion. If this is the first message of the discussion, we write "Nobody proposed a solution yet. Please provide the first one." instead of the most recent draft and agent memory.}
\end{figure}

%% file: prompts/solutionExtraction.tex
\begin{figure}[H]
\scalebox{0.5}{%
    \begin{tikzpicture}
    \node [draw, rectangle, rounded corners, fill=gray!20, text width=0.95\textwidth, inner sep=10pt] (block) {
        \begin{minipage}{\textwidth}
        \textbf{You are tasked with creating a final solution based on the given input and your previous response.} \\
        
        \textbf{Task:} \texttt{<instruction>} \\
        \textbf{Input:} \texttt{<example>} \\
        \textbf{Your previous response:} \texttt{<answer to extract solution from>} \\
        
        \textbf{Extract the final solution to the task from the provided text. Remove statements of agreement, disagreement, and explanations. Do not modify the text. Do not output any text besides the solution. If there is no solution provided, just copy the previous response.}
        \end{minipage}
    };
    \end{tikzpicture}
}
\caption{Prompt to extract the solution from an agent's answer.}
\end{figure}

%% file: prompts/voting.tex
\begin{figure}[H]
\scalebox{0.5}{%
    \begin{tikzpicture}
    \node [draw, rectangle, rounded corners, fill=gray!20, text width=0.95\textwidth, inner sep=10pt] (block) {
        \begin{minipage}{\textwidth}
        \textbf{Your role:} \texttt{<persona name>} (\texttt{<persona description>}) \\
        
        \textbf{You are tasked with voting for the best solution from the list provided below based on the given task.} \\
        \textbf{Task:} \texttt{<instruction>} \\
        \textbf{Question:} \texttt{<example>} \\
        \textbf{Additional Context: } \\

        \textbf{Here are the possible solutions:} \\
        \textbf{Solution A:} \texttt{<solution>} \\
        \textbf{Solution B:} \texttt{<solution>} \\
        \textbf{Solution C:} \texttt{<solution>} \\
        \end{minipage}
    };
    \end{tikzpicture}
}
\caption{Prompt to process the voting at the end of each turn. The number of solutions to vote for can vary depending on the proposals made by the agents.}
\end{figure}

%% file: prompts/personaGenerator.tex
\begin{figure}[H]
\scalebox{0.5}{%
    \begin{tikzpicture}
    \node [draw, rectangle, rounded corners, fill=gray!20, text width=0.95\textwidth, inner sep=10pt] (block) {
        \begin{minipage}{\textwidth}
        \textbf{When faced with a task, begin by identifying the participants who will contribute to solving the task. Provide role and description of the participants, describing their expertise or needs, formatted using the provided JSON schema. Generate one participant at a time, complementing the existing participants to foster a rich discussion.}\\
        \\
        \texttt{Example 1:}\\
        \texttt{Task: Explain the basics of machine learning to high school students.}\\
        \texttt{New Participant:}\\
        \texttt{\{"role": "Educator", "description": "An experienced teacher who simplifies complex topics for teenagers."\}}\\
        \\
        \texttt{Example 2:}\\
        \texttt{Task: Develop a new mobile app for tracking daily exercise.}\\
        \texttt{Already Generated Participants:}\\
        \texttt{\{"role": "Fitness Coach", "description": "A person that has high knowledge about sports and fitness."\}}\\
        \texttt{New Participant:}\\
        \texttt{\{"role": "Software Developer", "description": "A creative developer with experience in mobile applications and user interface design."\}}\\
        \\
        \texttt{Example 3:}\\
        \texttt{Task: Write a guide on how to cook Italian food for beginners.}\\
        \texttt{Already Generated Participants:}\\
        \texttt{\{"role": "Italian Native", "description": "An average home cook that lived in italy for 30 years."\}}\\
        \texttt{\{"role": "Food Scientist", "description": "An educated scientist that knows which flavor combinations result in the best taste."\}}\\
        \texttt{New Participant:}\\
        \texttt{\{"role": "Chef", "description": "A professional chef specializing in Italian cuisine who enjoys teaching cooking techniques."\}}\\
        \\
        \textbf{Now generate a participant to discuss the following task:} \\
        \textbf{Task:} \texttt{<instruction>} \\
        \textbf{Please use the follow the examples to generate a useful persona for the task!} \\
        \textbf{Only answer with the JSON for the next persona!} \\
        \textbf{Already Generated Participants:} \\
        \texttt{<list of already generated personas>}
        \end{minipage}
    };
    \end{tikzpicture}
}
\caption{Prompt for the automatic persona assignment. We generate the personas with three iterations of this prompt, adding one persona at a time that complements the ones previously generated.}
\end{figure}

%% file: prompts/llmJudge.tex
\begin{figure}[H]
\scalebox{0.5}{%
    \begin{tikzpicture}
    \node [draw, rectangle, rounded corners, fill=gray!20, text width=0.95\textwidth, inner sep=10pt] (block) {
        \begin{minipage}{\textwidth}
        \textbf{Please act as an impartial judge and evaluate the quality of the responses provided by two AI assistants to the user question displayed below. You should choose the assistant that follows the user\'s instructions and answers the user\'s question better.
        Your evaluation should consider factors such as the helpfulness, relevance, accuracy, depth, creativity, and level of detail of their responses. Avoid any position biases and ensure that the order in which the responses were presented does not influence your decision. Do not allow the length of the responses to influence your evaluation. Do not favor certain names of the assistants. Be as objective as possible.
        Please directly output your final verdict by strictly following this format: "[[A]]" if assistant A is better, "[[B]]" if assistant B is better.} \\

        \textbf{[User Question]} \\
        \texttt{<input>} \\

        \textbf{[The Start of Assistant A's Answer]} \\
        \texttt{<response of the previous turn>} \\
        \textbf{[The End of Assistant A's Answer]} \\

        \textbf{[The Start of Assistant B's Answer]} \\
        \texttt{<response of the current turn>} \\
        \textbf{[The End of Assistant B's Answer]} \\
        \end{minipage}
    };
    \end{tikzpicture}
}
\caption{Prompt for the LLM-as-a-judge.}
\end{figure}

%% file: prompts/policyFeedback.tex
\begin{figure}[H]
\scalebox{0.5}{%
    \begin{tikzpicture}
    \node [draw, rectangle, rounded corners, fill=gray!20, text width=0.95\textwidth, inner sep=10pt] (block) {
        \begin{minipage}{\textwidth}
        \textbf{The current discussion is going badly. Based on the others' contributions, give constructive feedback about how to improve the discussion habits. Be concise so that the other discussion participants can find a better solution.}\\

        \textbf{The following problematic error categories exist. If you identify them in the current discussion, they could help you to provide better feedback:}\\
        \textbf{Task Compliance:} \texttt{Off-topic, Bad instruction following}\\
        \textbf{Lack of Progress:} \texttt{Inefficiency, Redundancy, Circular discussion, Repetition, Unproductive disagreement}\\
        \textbf{Low-Quality Engagement:} \texttt{Poor collaboration, Minimal participation, Disjointed contribution, Ignorance}\\
        \textbf{Low-Quality Feedback:} \texttt{Excessive criticism, Excessive agreement, Self-contradictory feedback, Unhelpful feedback}\\
        \textbf{Lack of Clarity:} \texttt{Overanalysis, Overgeneralization, Insignificant changes}\\
        \textbf{Knowledge Gap:} \texttt{Assumptions, Lack of data, Hallucinated facts, Wrongly cited}\\
        \textbf{Logical Errors:} \texttt{Lack of common sense, Reasoning error}\\
        \textbf{Linguistic Errors:} \texttt{Fluency, Grammatical errors, False pronouns}\\
        \textbf{Other:} \texttt{Describe as explanation}\\

        \textbf{[The Start of Assistant A's Answer]} \\
        \texttt{<response of the previous turn>}\\
        \textbf{[The End of Assistant A's Answer]} \\

        \textbf{[The Start of Assistant B's Answer]} \\
        \texttt{<response of the current turn>}\\
        \textbf{[The End of Assistant B's Answer]} \\
        \end{minipage}
    };
    \end{tikzpicture}
}
\caption{Prompt for the policy feedback agent.}
\end{figure}

%% file: tables/datasets.tex
\begin{table*}[h]
\centering
\begin{tabular}{|p{3cm}|p{8cm}|p{2cm}r|}
\toprule
\textbf{Dataset} & \textbf{Description} & \textbf{Metrics} & \textbf{Samples} \\
\midrule
XSum \citep{NarayanCL18b} & Summarize a news article into a single sentence. & \makecell[tl]{BERTScore} & 386\tiny{$(\times3)$} \\
ETPC \citep{KovatchevMS18b} & Paraphrase a sentence based on a set of paraphrase types (e.g., addition/deletion, punctuation changes). & \makecell[tl]{BERTScore} & 361\tiny{$(\times3)$} \\
WMT19 (de-en) \citep{WikimediaFoundation19} & Translate a single sentence from English to German. & \makecell[tl]{BERTScore} & 341\tiny{$(\times3)$} \\
\midrule
StrategyQA \citep{GevaKSK21} & Multiple-choice questions that require strategic reasoning and planning to infer the correct answer. & Accuracy & 330\tiny{$(\times3)$} \\
WinoGrande \citep{SakaguchiBBC19} & Fill-in-the-blank task with binary options that require reasoning. & Accuracy & 295\tiny{$(\times3)$} \\
AQUA-RAT \citep{LingYDB17a} & Algebraic word problems with multiple-choice options. & Accuracy & 254\tiny{$(\times3)$} \\
\midrule
ETHICS \citep{HendrycksBBC23} & Multiple-choice benchmark for commonsense morality. & Accuracy & 351\tiny{$(\times3)$} \\
MMLU-Pro \citep{WangMZN24} & Adds more challenging examples to the MMLU dataset \citep{HendrycksBBZ21a}. & Accuracy & 373\tiny{$(\times3)$} \\
GPQA \citep{ReinHSP23a} & Google-proof multiple-choice questions written by experts from biology, physics, and chemistry. & Accuracy & 226\tiny{$(\times3)$} \\
\midrule
IFEval \citep{ZhouLMB23} & Tests instruction-following by variable prompts with 25 instruction types. & Accuracy & 541\tiny{$(\times3)$} \\
\bottomrule
\end{tabular}
\caption{Datasets with the number of samples used in the experiments extracted randomly by a 95\% confidence interval and a 5\% margin of error ($MoE$), conservatively assuming a sample proportion $p=0.5$. We randomly sample three times from each dataset and report the standard deviations in metric scores between the five runs. The top three datasets are generative tasks, the middle three datasets are reasoning-heavy tasks, and the bottom three tasks are knowledge-intensive tasks. IFEval concerns instruction-following, coming with a separate benchmark script. Their "strict" accuracy is used in this work.}
\label{tab:datasets}
\end{table*}

%% file: examples/tex/lack_of_progress.tex
\definecolor{lightyellow}{RGB}{255, 255, 204}

\begin{minipage}{\textwidth}
\scalebox{0.9}{%
\begin{tcolorbox}[title=Sample Information, colframe=black!10, coltitle=black, fonttitle=\bfseries, boxrule=0.5mm, width=\textwidth, fontupper=\small]

\textbf{Task: }
Translate the provided text from German to English.

\textbf{Input: }
"Eine solche Aussage zu treffen wäre arrogant und überheblich".

\textbf{Hidden reference:}
"To give such a statement would be arrogant and presumptuous."

\end{tcolorbox}
}

\scalebox{0.9}{%
\begin{tcolorbox}[breakable, title=Turn 3, colframe=black!10, coltitle=black, fonttitle=\bfseries, boxrule=0.5mm, width=\textwidth, fontupper=\small]

\begin{tcolorbox}[title=Professional Translator, colback=lightyellow, colframe=orange!30, coltitle=black, fonttitle=\bfseries, boxrule=0.5mm, width=\textwidth]
\begin{minipage}{0.1\textwidth}
    \centering
    \hspace{-5mm}\includegraphics[width=0.9\textwidth]{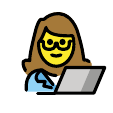} \\
    \hspace{-5mm}\includegraphics[width=0.5\textwidth]{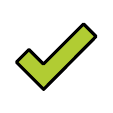}
\end{minipage}
\begin{minipage}{0.9\textwidth}
    [AGREE] After thoroughly examining the current solution and the discussion that led to it, I agree that the translation "To make such a statement would be arrogant and condescending" is accurate and idiomatic. The breakdown of the sentence and consideration of the cultural expressions were thorough, and the choice of "condescending" to translate "überheblich" effectively conveys the sense of superiority and belittling attitude present in the original sentence. The revised sentence structure "To make such a statement" is also a common English idiomatic expression that flows naturally. Moreover, the agreement from multiple perspectives, including a Native English Speaker, a Professional Translator, and a Native German Speaker, provides confidence in the solution's validity. The lack of suggestions for improvement from any of the participants in the discussion also supports the notion that the current solution is satisfactory. Given the thorough analysis and the consensus among the participants, I conclude that the current solution is indeed accurate and idiomatic, and I see no need for further refinement. However, it's worth noting that minor tweaks may be considered for specific contexts or audiences, but as a general translation, the current solution is sufficient. Overall, the current solution effectively captures the essence of the original German sentence, and I don't see any areas for improvement.
\end{minipage}
\end{tcolorbox}

\begin{tcolorbox}[title=Native English Speaker, colback=lightyellow, colframe=orange!30, coltitle=black, fonttitle=\bfseries, boxrule=0.5mm, width=\textwidth]
\begin{minipage}{0.1\textwidth}
    \centering
    \hspace{-5mm}\includegraphics[width=0.9\textwidth]{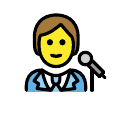} \\
    \hspace{-5mm}\includegraphics[width=0.5\textwidth]{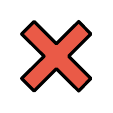}
\end{minipage}
\begin{minipage}{0.9\textwidth}
    [DISAGREE] While the current solution, "To make such a statement would be arrogant and condescending," is a good translation, I believe it can be slightly improved for more formal or written contexts. The phrase "such a statement" is a bit general and could be rephrased to more accurately capture the essence of the original sentence. I propose rephrasing it to "such an assertion" or "such a claim." This change adds a bit more formal tone and precision to the translation. Additionally, I suggest using "presumptuous" instead of "arrogant." While "arrogant" is a good translation, "presumptuous" more accurately conveys the sense of overstepping or overconfidence that is present in the original sentence. Here's my revised solution: \textbf{"To make such a claim would be presumptuous and condescending."} This revised solution maintains the accuracy and idiomacity of the original while providing a more precise and formal translation. However, I must note that the original solution is still suitable for general contexts, and my revised solution is more geared towards formal or written contexts.
\end{minipage}
\end{tcolorbox}

\begin{tcolorbox}[title=Native German Speaker, colback=lightyellow, colframe=orange!30, coltitle=black, fonttitle=\bfseries, boxrule=0.5mm, width=\textwidth]
\begin{minipage}{0.1\textwidth}
    \centering
    \hspace{-5mm}\includegraphics[width=0.9\textwidth]{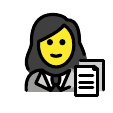} \\
    \hspace{-5mm}\includegraphics[width=0.5\textwidth]{examples/cross.pdf}
\end{minipage}
\begin{minipage}{0.9\textwidth}
    [DISAGREE] I agree with the Native English Speaker's suggestion to refine the current solution for more formal or written contexts. The phrase "such a statement" is indeed quite general, and using "such a claim" instead adds a bit more precision and formality to the translation. I also agree that using "presumptuous" instead of "arrogant" better conveys the sense of overstepping or overconfidence that is present in the original sentence. "Presumptuous" implies a sense of assuming or taking something for granted without sufficient evidence or justification, which fits well with the German phrase "überheblich". The revised solution proposed by the Native English Speaker, "To make such a claim would be presumptuous and condescending", is an improvement over the current solution. It maintains the accuracy and idiomacity of the original while providing a more precise and formal translation. However, I would like to propose a slight refinement to this revised solution. Instead of using "such a claim", I suggest using "such an assertion". "Assertion" implies a stronger sense of stating or declaring something as fact, which fits well with the German phrase "Aussage". Here's my revised solution: \textbf{"To make such an assertion would be presumptuous and condescending"}. This revised solution provides a more precise and formal translation, while maintaining the accuracy and idiomacity of the original sentence.
\end{minipage}
\end{tcolorbox}

\begin{tcolorbox}[coltitle=black, fonttitle=\bfseries, boxrule=0.5mm, width=\textwidth]
\textbf{Voted solution of this turn:}\\ "To make such a statement would be arrogant and condescending." \hfill \texttt{BLEU: 56.59}
\end{tcolorbox}

\end{tcolorbox}
}
\end{minipage}

\scalebox{0.9}{%
\begin{tcolorbox}[title=Turn 4, colframe=black!10, coltitle=black, fonttitle=\bfseries, boxrule=0.5mm, width=\textwidth, fontupper=\small]

\begin{tcolorbox}[title=Professional Translator, colback=lightyellow, colframe=orange!30, coltitle=black, fonttitle=\bfseries, boxrule=0.5mm, width=\textwidth]
\begin{minipage}{0.1\textwidth}
    \centering
    \hspace{-5mm}\includegraphics[width=0.9\textwidth]{examples/ethicist.pdf} \\
    \hspace{-5mm}\includegraphics[width=0.5\textwidth]{examples/cross.pdf}
\end{minipage}
\begin{minipage}{0.9\textwidth}
    [DISAGREE] After considering the feedback from the Native English Speaker and the Native German Speaker, I agree that the current solution can be improved for more formal or written contexts. The suggestions to use "such a claim" or "such an assertion" instead of "such a statement" add more precision and formality to the translation. I prefer "such an assertion" because it implies a stronger sense of stating or declaring something as fact, which fits well with the German phrase "Aussage". The suggestion to use "presumptuous" instead of "arrogant" also improves the translation. "Presumptuous" more accurately conveys the sense of overstepping or overconfidence that is present in the original sentence. Based on these suggestions, I propose the following improved solution: \textbf{"To make such an assertion would be presumptuous and condescending."} This revised solution maintains the accuracy and idiomacity of the original while providing a more precise and formal translation. It is suitable for formal or written contexts where a more precise and nuanced translation is required. Here's a breakdown of the changes: * "Such a statement" -> "such an assertion" (more precise and formal) * "Arrogant" -> "presumptuous" (more accurately conveys the sense of overstepping or overconfidence) Overall, the revised solution is an improvement over the current solution, and I believe it provides a more accurate and idiomatic translation of the original German sentence.
\end{minipage}
\end{tcolorbox}

\begin{tcolorbox}[title=Native English Speaker, colback=lightyellow, colframe=orange!30, coltitle=black, fonttitle=\bfseries, boxrule=0.5mm, width=\textwidth]
\begin{minipage}{0.1\textwidth}
    \centering
    \hspace{-5mm}\includegraphics[width=0.9\textwidth]{examples/politician.pdf} \\
    \hspace{-5mm}\includegraphics[width=0.5\textwidth]{examples/check.pdf}
\end{minipage}
\begin{minipage}{0.9\textwidth}
    [AGREE] After considering the feedback and suggestions from all parties involved, I agree with the revised solution: "To make such an assertion would be presumptuous and condescending." This revised solution is an improvement over the current solution, and I believe it provides a more accurate and idiomatic translation of the original German sentence. The changes made to the original solution, specifically using "such an assertion" instead of "such a statement" and "presumptuous" instead of "arrogant," significantly improve the translation. "Such an assertion" adds more precision and formality, while "presumptuous" more accurately conveys the sense of overstepping or overconfidence. I also appreciate the thorough breakdown of the changes made to the solution. The explained changes help to ensure that all parties are in agreement and understand the reasoning behind the revised translation. As the revised solution has been carefully crafted and refined through the collaborative efforts of multiple perspectives, including a Native German Speaker, a Professional Translator, and a Native English Speaker, I believe that it is the most accurate and idiomatic translation of the original German sentence. Therefore, I conclude that no further refinement is needed, and I support the revised solution: "To make such an assertion would be presumptuous and condescending."
\end{minipage}
\end{tcolorbox}

\begin{tcolorbox}[title=Native German Speaker, colback=lightyellow, colframe=orange!30, coltitle=black, fonttitle=\bfseries, boxrule=0.5mm, width=\textwidth]
\begin{minipage}{0.1\textwidth}
    \centering
    \hspace{-5mm}\includegraphics[width=0.9\textwidth]{examples/lawyer.pdf} \\
    \hspace{-5mm}\includegraphics[width=0.5\textwidth]{examples/check.pdf}
\end{minipage}
\begin{minipage}{0.9\textwidth}
    [AGREE] I agree with the revised solution: "To make such an assertion would be presumptuous and condescending." The changes made to the original solution, specifically using "such an assertion" instead of "such a statement" and "presumptuous" instead of "arrogant," significantly improve the translation. "Such an assertion" adds more precision and formality, while "presumptuous" more accurately conveys the sense of overstepping or overconfidence. I also appreciate the thorough breakdown of the changes made to the solution. The explained changes help to ensure that all parties are in agreement and understand the reasoning behind the revised translation. As the revised solution has been carefully crafted and refined through the collaborative efforts of multiple perspectives, including a Native German Speaker, a Professional Translator, and a Native English Speaker, I believe that it is the most accurate and idiomatic translation of the original German sentence. Therefore, I conclude that no further refinement is needed, and I support the revised solution: "To make such an assertion would be presumptuous and condescending." This solution effectively captures the nuances of the original German sentence, and I see no need for further improvement.
\end{minipage}
\end{tcolorbox}

\begin{tcolorbox}[coltitle=black, fonttitle=\bfseries, boxrule=0.5mm, width=\textwidth]
\textbf{Voted solution of this turn:}\\ To make such an assertion would be presumptuous and condescending. \hfill \texttt{BLEU: 0.0}
\end{tcolorbox}

\end{tcolorbox}
}

\label{example:lack_of_progress}

%% file: examples/tex/low_quality_feedback.tex
\definecolor{lightyellow}{RGB}{255, 255, 204}

\begin{minipage}{\textwidth}
\scalebox{0.9}{%
\begin{tcolorbox}[title=Sample Information, colframe=black!10, coltitle=black, fonttitle=\bfseries, boxrule=0.5mm, width=\textwidth, fontupper=\small]

\textbf{Task: }
Follow the instruction.

\textbf{Input: }
What is the name of the actor who played Gandalf in Lord of the Rings? First repeat the question above without change of words, then give your answer.

\textbf{Hidden reference:}
What is the name of the actor who played Gandalf in Lord of the Rings?\\
The answer is Ian McKellen.

\end{tcolorbox}
}

\scalebox{0.9}{%
\begin{tcolorbox}[breakable, title=Turn 1, colframe=black!10, coltitle=black, fonttitle=\bfseries, boxrule=0.5mm, width=\textwidth, fontupper=\small]

\begin{tcolorbox}[title=Cinema Expert, colback=lightyellow, colframe=orange!30, coltitle=black, fonttitle=\bfseries, boxrule=0.5mm, width=\textwidth]
\begin{minipage}{0.1\textwidth}
    \centering
    \hspace{-5mm}\includegraphics[width=0.9\textwidth]{examples/ethicist.pdf} \\
    \hspace{-5mm}\includegraphics[width=0.5\textwidth]{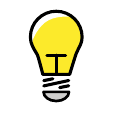}
\end{minipage}
\begin{minipage}{0.9\textwidth}
    To solve this task, I will follow the provided instructions. Here's my step-by-step solution: 1. First, I will repeat the question exactly as it was given:  \textbf{What is the name of the actor who played Gandalf in Lord of the Rings?} 2. Next, I will provide the answer based on my knowledge as a Cinema Expert: \textbf{The answer is Ian McKellen.}
\end{minipage}
\end{tcolorbox}

\begin{tcolorbox}[title=Pop Culture Enthusiast, colback=lightyellow, colframe=orange!30, coltitle=black, fonttitle=\bfseries, boxrule=0.5mm, width=\textwidth]
\begin{minipage}{0.1\textwidth}
    \centering
    \hspace{-5mm}\includegraphics[width=0.9\textwidth]{examples/politician.pdf} \\
    \hspace{-5mm}\includegraphics[width=0.5\textwidth]{examples/check.pdf}
\end{minipage}
\begin{minipage}{0.9\textwidth}
    [AGREE] I agree with the current solution. The Cinema Expert has correctly followed the instructions provided. They have repeated the question exactly as it was given in step 1 and provided the correct answer in step 2. Ian McKellen is indeed the actor who played Gandalf in the Lord of the Rings film series. However, as a Pop Culture Enthusiast, I would like to add some additional information that may not be necessary to solve the task but could be interesting for fans of the series.  Ian McKellen's portrayal of Gandalf is widely praised and is often considered one of the most iconic roles in the series. He also played the same character in the Hobbit trilogy, which serves as a prequel to the Lord of the Rings.
\end{minipage}
\end{tcolorbox}

\begin{tcolorbox}[title=Movie Database Manager, colback=lightyellow, colframe=orange!30, coltitle=black, fonttitle=\bfseries, boxrule=0.5mm, width=\textwidth]
\begin{minipage}{0.1\textwidth}
    \centering
    \hspace{-5mm}\includegraphics[width=0.9\textwidth]{examples/lawyer.pdf} \\
    \hspace{-5mm}\includegraphics[width=0.5\textwidth]{examples/check.pdf}
\end{minipage}
\begin{minipage}{0.9\textwidth}
    As the Movie Database Manager, I will review the current solution and provide my assessment.  [AGREE]  I agree with the current solution. It correctly follows the instructions provided by repeating the question exactly as it was given and providing the correct answer. The additional information provided by the Pop Culture Enthusiast is interesting and adds context to Ian McKellen's role as Gandalf, but it is not necessary for solving the task.  The current solution is simple, accurate, and effectively addresses the task at hand. As the Movie Database Manager, I have verified that Ian McKellen is indeed the actor who played Gandalf in the Lord of the Rings film series, and this information is consistent with my film database.
\end{minipage}
\end{tcolorbox}

\begin{tcolorbox}[coltitle=black, fonttitle=\bfseries, boxrule=0.5mm, width=\textwidth]
\textbf{Voted solution of this turn:}\\ What is the name of the actor who played Gandalf in Lord of the Rings? \\
The answer is Ian McKellen. \hfill \texttt{Correct: True}
\end{tcolorbox}

\end{tcolorbox}
}
\end{minipage}

\scalebox{0.9}{%
\begin{tcolorbox}[title=Turn 2, colframe=black!10, coltitle=black, fonttitle=\bfseries, boxrule=0.5mm, width=\textwidth, fontupper=\small]

\begin{tcolorbox}[title=Cinema Expert, colback=lightyellow, colframe=orange!30, coltitle=black, fonttitle=\bfseries, boxrule=0.5mm, width=\textwidth]
\begin{minipage}{0.1\textwidth}
    \centering
    \hspace{-5mm}\includegraphics[width=0.9\textwidth]{examples/ethicist.pdf} \\
    \hspace{-5mm}\includegraphics[width=0.5\textwidth]{examples/check.pdf}
\end{minipage}
\begin{minipage}{0.9\textwidth}
    [AGREE] I agree with the current solution. The Cinema Expert correctly followed the instructions provided by repeating the question exactly as it was given in step 1 and providing the correct answer in step 2. Ian McKellen is indeed the actor who played Gandalf in the Lord of the Rings film series. The Pop Culture Enthusiast provided additional information that is interesting and adds context to Ian McKellen's role as Gandalf, and the Movie Database Manager verified the accuracy of this information. At this point, the solution seems complete and effective, and there is no need for further improvements.
\end{minipage}
\end{tcolorbox}

\begin{tcolorbox}[title=Pop Culture Enthusiast, colback=lightyellow, colframe=orange!30, coltitle=black, fonttitle=\bfseries, boxrule=0.5mm, width=\textwidth]
\begin{minipage}{0.1\textwidth}
    \centering
    \hspace{-5mm}\includegraphics[width=0.9\textwidth]{examples/politician.pdf} \\
    \hspace{-5mm}\includegraphics[width=0.5\textwidth]{examples/cross.pdf}
\end{minipage}
\begin{minipage}{0.9\textwidth}
    [DISAGREE] While the current solution is indeed simple, accurate, and effectively addresses the task, I believe it can be improved by including a minor clarification that would make the answer more comprehensive. As the Pop Culture Enthusiast, I noticed that Ian McKellen played two different versions of Gandalf in the Lord of the Rings and the Hobbit film trilogies: Gandalf the Grey in the first two movies of the Lord of the Rings trilogy and some of the Hobbit trilogy, and Gandalf the White in the rest of the Lord of the Rings trilogy and some of the Hobbit trilogy. Although this clarification is not necessary for solving the task, including it might make the answer more informative and interesting for fans of the series. Here's my improved solution: \textbf{1. Repeated Question: What is the name of the actor who played Gandalf in Lord of the Rings? 2. Answer: The answer is Ian McKellen. He played two different versions of Gandalf in the Lord of the Rings and the Hobbit film trilogies: Gandalf the Grey and Gandalf the White, due to the events within the story.} By adding this minor clarification, the improved solution provides more context about Ian McKellen's portrayal of Gandalf, without straying away from the main question and making it unnecessarily complicated. Please share your thoughts, and we can discuss whether this improvement is necessary or not.
\end{minipage}
\end{tcolorbox}

\begin{tcolorbox}[title=Movie Database Manager, colback=lightyellow, colframe=orange!30, coltitle=black, fonttitle=\bfseries, boxrule=0.5mm, width=\textwidth]
\begin{minipage}{0.1\textwidth}
    \centering
    \hspace{-5mm}\includegraphics[width=0.9\textwidth]{examples/lawyer.pdf} \\
    \hspace{-5mm}\includegraphics[width=0.5\textwidth]{examples/check.pdf}
\end{minipage}
\begin{minipage}{0.9\textwidth}
    Movie Database Manager: [AGREE]  I agree with the improved solution provided by the Pop Culture Enthusiast. Although the original solution was simple and accurate, the additional clarification provides more context about Ian McKellen's portrayal of Gandalf. This clarification is intriguing for fans of the series and can be beneficial for those who are interested in the nuances of the character.  The improved solution does not stray away from the main question and is still concise and easy to understand. The additional information is relevant and does not make the answer unnecessarily complicated.  In film databases, it is common to note different versions or iterations of a character played by the same actor. Therefore, I believe the improved solution is a good representation of how Ian McKellen's role as Gandalf can be accurately documented.  The improved solution is:  1. Repeated Question: What is the name of the actor who played Gandalf in Lord of the Rings? 2. Answer: The answer is Ian McKellen. He played two different versions of Gandalf in the Lord of the Rings and the Hobbit film trilogies: Gandalf the Grey and Gandalf the White, due to the events within the story.  This improved solution has been verified for accuracy, and I believe it is an effective and comprehensive answer to the original question.
\end{minipage}
\end{tcolorbox}

\begin{tcolorbox}[coltitle=black, fonttitle=\bfseries, boxrule=0.5mm, width=\textwidth]
\textbf{Voted solution of this turn:}\\ 1. Repeated Question: \\What is the name of the actor who played Gandalf in Lord of the Rings? \\\\2. Answer: \\The answer is Ian McKellen. He played two different versions of Gandalf in the Lord of the Rings and the Hobbit film trilogies: Gandalf the Grey and Gandalf the White, due to the events within the story. \hfill \texttt{Correct: False}
\end{tcolorbox}

\end{tcolorbox}
}

\label{example:low_quality_feedback}

%% file: examples/tex/low_quality_engagement.tex
\definecolor{lightyellow}{RGB}{255, 255, 204}

\begin{minipage}{\textwidth}
\scalebox{0.9}{%
\begin{tcolorbox}[title=Sample Information, colframe=black!10, coltitle=black, fonttitle=\bfseries, boxrule=0.5mm, width=\textwidth, fontupper=\small]

\textbf{Task: }
Answer the provided question by choosing one option. Include the letter corresponding to your answer in the solution.

\textbf{Input: }
Experiencing intense feelings of shame and guilt upon admitting to a counselor that a family problem exists is common among traditional A) Eastern European-Americans B) Asian-Americans
C) African-Americans D) Native North Americans
E) European-Americans F) Hispanics G) Caribbean-Americans

\textbf{Hidden reference:}
B

\end{tcolorbox}
}

\scalebox{0.9}{%
\begin{tcolorbox}[breakable, title=Turn 1, colframe=black!10, coltitle=black, fonttitle=\bfseries, boxrule=0.5mm, width=\textwidth, fontupper=\small]

\begin{tcolorbox}[title=Cross-Cultural Psychologist, colback=lightyellow, colframe=orange!30, coltitle=black, fonttitle=\bfseries, boxrule=0.5mm, width=\textwidth]
\begin{minipage}{0.1\textwidth}
    \centering
    \hspace{-5mm}\includegraphics[width=0.9\textwidth]{examples/ethicist.pdf} \\
    \hspace{-5mm}\includegraphics[width=0.5\textwidth]{examples/bulp.pdf}
\end{minipage}
\begin{minipage}{0.9\textwidth}
    [DISAGREE] While the emphasis on "saving face" in Asian-American cultures is indeed a crucial factor in the intensity of feelings of shame and guilt when admitting to family problems, the discussion should consider other perspectives more thoroughly. However, I still believe that among the provided options, **Asian-Americans** have a more intense emphasis on "saving face" and the potential impact of shame in the context of family-related admissions. Considering the provided explanation that in collective over individual, hierarchical relationships and pragmatic communication styles often characterize collectivist cultures, it doesn't necessarily change the initial solution since Asian-American culture is considered a collectivist culture, which aligns with these characteristics. The repetitive mention of collective over individual, hierarchical relationships, and pragmatic communication styles doesn't provide any new evidence or an alternative choice that would challenge the current solution. Therefore, although there are some disagreements with the initial approach, after thorough analysis, I still believe that among the provided options: **Solution: \textbf{B}** Asian-Americans are a more accurate answer given the information provided.
\end{minipage}
\end{tcolorbox}

\begin{tcolorbox}[title=Counselor, colback=lightyellow, colframe=orange!30, coltitle=black, fonttitle=\bfseries, boxrule=0.5mm, width=\textwidth]
\begin{minipage}{0.1\textwidth}
    \centering
    \hspace{-5mm}\includegraphics[width=0.9\textwidth]{examples/politician.pdf} \\
    \hspace{-5mm}\includegraphics[width=0.5\textwidth]{examples/cross.pdf}
\end{minipage}
\begin{minipage}{0.9\textwidth}
    [DISAGREE] While the initial explanation and Cross-Cultural Psychologist's response present a clear analysis of the emphasis on "saving face" in Asian-American cultures and its impact on feelings of shame and guilt when admitting to family problems, there's a need to refine the solution. The Anthropologist's point about the definition of 'Traditional' and its implications on collective, hierarchical relationships, and pragmatic communication styles in collectivist cultures provides insightful context. However, the repetition of this concept doesn't lead to an alternative explanation. Considering the options provided and the explanations, I agree that Asian-Americans are likely to experience intense feelings of shame and guilt when admitting to family problems. Nonetheless, the Anthropologist's observation on the role of collectivist cultures should be considered. However, instead of providing an improved solution with an alternative option, I suggest refining the current solution by recognizing that cultures other than Asian-Americans (like some Hispanic cultures) might share similarities with the described characteristics of collectivist cultures. The explanation, however, points out that between Hispanics and Asian-Americans, the impact of shame might indeed be more nuanced in the context of family-related admissions. So based on the given options, the explanation, and the analysis of all points raised, I do not find any concrete reason to change the solution. So, refining and accepting the solution:  **Solution: \textbf{B}**
\end{minipage}
\end{tcolorbox}

\begin{tcolorbox}[title=Anthropologist, colback=lightyellow, colframe=orange!30, coltitle=black, fonttitle=\bfseries, boxrule=0.5mm, width=\textwidth]
\begin{minipage}{0.1\textwidth}
    \centering
    \hspace{-5mm}\includegraphics[width=0.9\textwidth]{examples/lawyer.pdf} \\
    \hspace{-5mm}\includegraphics[width=0.5\textwidth]{examples/check.pdf}
\end{minipage}
\begin{minipage}{0.9\textwidth}
    [AGREE] After analyzing the discussion and considering the various perspectives, I agree with the current solution: B (Asian-Americans). While the Anthropologist's repeated mention of collective, hierarchical relationships, and pragmatic communication styles doesn't necessarily challenge the solution, it does provide additional context that supports the notion that collectivist cultures, such as Asian-American culture, may be more likely to experience intense feelings of shame and guilt when admitting to family problems. The Cross-Cultural Psychologist's acknowledgment of the importance of "saving face" in Asian-American cultures and its impact on the intensity of feelings of shame and guilt when admitting to family problems aligns with the explanation provided. The Counselor's suggestion to refine the solution by recognizing that other collectivist cultures, such as some Hispanic cultures, might share similarities with the described characteristics is a valuable point. However, the explanation does note that between Hispanics and Asian-Americans, the impact of shame might be more nuanced in the context of family-related admissions. Given the analysis and the explanations provided, I don't see any concrete reason to change the solution. The current solution, B (Asian-Americans), remains the most accurate answer given the information provided. Therefore, I agree with the current solution. **Solution: B**
\end{minipage}
\end{tcolorbox}

\begin{tcolorbox}[coltitle=black, fonttitle=\bfseries, boxrule=0.5mm, width=\textwidth]
\textbf{Voted solution of this turn:}\\ B) Asian-Americans \hfill \texttt{Correct: True}
\end{tcolorbox}

\end{tcolorbox}
}
\end{minipage}

\scalebox{0.9}{%
\begin{tcolorbox}[title=Turn 2, colframe=black!10, coltitle=black, fonttitle=\bfseries, boxrule=0.5mm, width=\textwidth, fontupper=\small]

\begin{tcolorbox}[title=Cross-Cultural Psychologist, colback=lightyellow, colframe=orange!30, coltitle=black, fonttitle=\bfseries, boxrule=0.5mm, width=\textwidth]
\begin{minipage}{0.1\textwidth}
    \centering
    \hspace{-5mm}\includegraphics[width=0.9\textwidth]{examples/ethicist.pdf} \\
    \hspace{-5mm}\includegraphics[width=0.5\textwidth]{examples/check.pdf}
\end{minipage}
\begin{minipage}{0.9\textwidth}
    [AGREE] After analyzing the discussion, considering the various perspectives, and evaluating the explanation provided, I believe the current solution is the most accurate answer given the information. The emphasis on "saving face" in Asian-American cultures, the impact of shame and guilt when admitting to family problems, and the characteristics of collectivist cultures all support the conclusion that Asian-Americans are more likely to experience intense feelings of shame and guilt in the context of family-related admissions. While other collectivist cultures, such as some Hispanic cultures, might share similarities with the described characteristics, the explanation highlights the nuances between these cultures in the context of family-related admissions. Therefore, the current solution remains the most accurate answer. **Solution: B** Since all parties have agreed with the solution and there are no objections or alternative explanations that would change the solution, the current solution is confirmed. The discussion has refined the solution by recognizing the shared characteristics of collectivist cultures, but ultimately, the current solution is the most accurate answer given the information provided.
\end{minipage}
\end{tcolorbox}

\begin{tcolorbox}[title=Counselor, colback=lightyellow, colframe=orange!30, coltitle=black, fonttitle=\bfseries, boxrule=0.5mm, width=\textwidth]
\begin{minipage}{0.1\textwidth}
    \centering
    \hspace{-5mm}\includegraphics[width=0.9\textwidth]{examples/politician.pdf} \\
    \hspace{-5mm}\includegraphics[width=0.5\textwidth]{examples/check.pdf}
\end{minipage}
\begin{minipage}{0.9\textwidth}
    [AGREE] Given the thorough analysis and agreement from all parties involved in the discussion, the current solution, B (Asian-Americans), is the most accurate answer based on the information provided. The discussion has effectively refined the solution by considering various perspectives, recognizing the importance of "saving face" in Asian-American cultures, and acknowledging the shared characteristics of collectivist cultures. While some nuances have been highlighted, such as the potential similarities between Asian-Americans and other collectivist cultures (e.g., some Hispanic cultures), the explanation has demonstrated that the impact of shame and guilt when admitting to family problems might be more pronounced in Asian-American cultures. Therefore, I agree with the current solution, and no improvement is needed. **Solution: B**
\end{minipage}
\end{tcolorbox}

\begin{tcolorbox}[title=Anthropologist, colback=lightyellow, colframe=orange!30, coltitle=black, fonttitle=\bfseries, boxrule=0.5mm, width=\textwidth]
\begin{minipage}{0.1\textwidth}
    \centering
    \hspace{-5mm}\includegraphics[width=0.9\textwidth]{examples/lawyer.pdf} \\
    \hspace{-5mm}\includegraphics[width=0.5\textwidth]{examples/cross.pdf}
\end{minipage}
\begin{minipage}{0.9\textwidth}
    [DISAGREE] While the current solution, B (Asian-Americans), has been thoroughly discussed and agreed upon by all parties, I would like to revisit the analysis and provide an alternative perspective. Upon re-examining the explanation and considering the characteristics of collectivist cultures, I believe that another option might be equally or even more plausible. **Hispanic culture**, especially in traditional settings, places a strong emphasis on family and social harmony. The concept of "respeto" (respect) and "orgullo" (family pride) can be equally influential in shaping the reactions of Hispanic individuals when admitting to family problems. In Hispanic cultures, the concept of "\u2019en familia no se mete" (don't meddle in family business) can lead to a strong sense of shame and guilt when family problems are disclosed. This cultural dynamic might be just as intense as the "saving face" concept in Asian-American cultures. Given this alternative perspective, I suggest considering Hispanic culture as a viable solution. **Improved Solution: \textbf{F}** Hispanics
\end{minipage}
\end{tcolorbox}

\begin{tcolorbox}[coltitle=black, fonttitle=\bfseries, boxrule=0.5mm, width=\textwidth]
\textbf{Voted solution of this turn:}\\ F) Hispanics \hfill \texttt{Correct: False}
\end{tcolorbox}

\end{tcolorbox}
}

\label{example:low_quality_engagement}

%% file: examples/tex/lack_of_clarity.tex
\definecolor{lightyellow}{RGB}{255, 255, 204}

\begin{minipage}{\textwidth}
\scalebox{0.9}{%
\begin{tcolorbox}[title=Sample Information, colframe=black!10, coltitle=black, fonttitle=\bfseries, boxrule=0.5mm, width=\textwidth, fontupper=\small]

\textbf{Task: }
Answer the provided question by choosing one option. Include the letter corresponding to your answer in the solution.

\textbf{Input: }
A passage Jane Eyre is as follows. This was all the account I got from Mrs. Fairfax of her employer and mine. There are people who seem to have no notion of sketching a character, or observing and describing salient points, either in persons or things: the good lady evidently belonged to this class; my queries puzzled, but did not draw her out. Mr. Rochester was Mr. Rochester in her eyes, a gentleman, a landed proprietor — nothing more: she inquired and searched no further, and evidently wondered at my wish to gain a more definite notion of his identity. Mrs. Fairfax differs from the speaker in that Mrs. Fairfax A) has a more critical view of people's character B) is more willing to take people at face value C) is less skeptical about people's intentions D) has a more positive opinion of Mr. Rochester E) has more interest in the complexities of people's personalities F) is more reserved in her observations G) judges people by their social station in life

\textbf{Hidden reference:}
B

\end{tcolorbox}
}

\scalebox{0.9}{%
\begin{tcolorbox}[breakable, title=Turn 1, colframe=black!10, coltitle=black, fonttitle=\bfseries, boxrule=0.5mm, width=\textwidth, fontupper=\small]

\begin{tcolorbox}[title=Literary Analyst, colback=lightyellow, colframe=orange!30, coltitle=black, fonttitle=\bfseries, boxrule=0.5mm, width=\textwidth]
\begin{minipage}{0.1\textwidth}
    \centering
    \hspace{-5mm}\includegraphics[width=0.9\textwidth]{examples/ethicist.pdf} \\
    \hspace{-5mm}\includegraphics[width=0.5\textwidth]{examples/bulp.pdf}
\end{minipage}
\begin{minipage}{0.9\textwidth}
    To solve this task, let's break it down step by step: 1. **Read the passage**: We have already read the passage from Jane Eyre. 2. **Understand the task**: We need to choose an option that describes how Mrs. Fairfax differs from the speaker. 3. **Analyze the speaker's approach**: The speaker seems inquisitive and wants to gain a more definite notion of Mr. Rochester's identity, asking Mrs. Fairfax more questions. 4. **Analyze Mrs. Fairfax's approach**: Mrs. Fairfax, on the other hand, sees Mr. Rochester as just a gentleman and a landed proprietor. She doesn't appear to want to know more about him and seems puzzled by the speaker's questions. 5. **Compare and contrast**: Given the two approaches, we can see that the speaker is more inquisitive and analytical, while Mrs. Fairfax is more accepting and doesn't probe deeper. 6. **Identify the difference**: The main difference between the speaker and Mrs. Fairfax is that Mrs. Fairfax is more willing to accept people at face value, without questioning or seeking more information. 7. **Choose the correct option**: Based on this analysis, the correct answer is **B) is more willing to take people at face value**. Solution: \textbf{B}
\end{minipage}
\end{tcolorbox}

\begin{tcolorbox}[title=Reluctant Reader, colback=lightyellow, colframe=orange!30, coltitle=black, fonttitle=\bfseries, boxrule=0.5mm, width=\textwidth]
\begin{minipage}{0.1\textwidth}
    \centering
    \hspace{-5mm}\includegraphics[width=0.9\textwidth]{examples/politician.pdf} \\
    \hspace{-5mm}\includegraphics[width=0.5\textwidth]{examples/check.pdf}
\end{minipage}
\begin{minipage}{0.9\textwidth}
    [AGREE] The steps you've outlined for analysis are clear and help break down the task into manageable parts. Your conclusion that Mrs. Fairfax is more willing to take people at face value aligns with the evidence presented in the passage. Your analysis effectively highlights the contrast between the speaker's inquisitive nature and Mrs. Fairfax's more accepting approach. By demonstrating that Mrs. Fairfax sees Mr. Rochester as simply a gentleman and a landed proprietor without seeking more information, you make a strong case for option B. The steps you've provided offer a good framework for approaching literary analysis tasks, and your application of these steps to the given task is sound. Therefore, I agree with the current solution and do not see a need for improvement in this case.
\end{minipage}
\end{tcolorbox}

\begin{tcolorbox}[title=Language Expert, colback=lightyellow, colframe=orange!30, coltitle=black, fonttitle=\bfseries, boxrule=0.5mm, width=\textwidth]
\begin{minipage}{0.1\textwidth}
    \centering
    \hspace{-5mm}\includegraphics[width=0.9\textwidth]{examples/lawyer.pdf} \\
    \hspace{-5mm}\includegraphics[width=0.5\textwidth]{examples/check.pdf}
\end{minipage}
\begin{minipage}{0.9\textwidth}
    [AGREE] I agree with the current solution, B) is more willing to take people at face value. The analysis provided by the Literary Analyst is thorough and accurately captures the difference between Mrs. Fairfax and the speaker. The steps outlined by the Literary Analyst are a great framework for approaching literary analysis tasks. They demonstrate a clear understanding of the task, a thorough reading of the passage, and a thoughtful analysis of the characters' approaches. The conclusion drawn from the analysis, that Mrs. Fairfax is more willing to take people at face value, is well-supported by the evidence in the passage. The Reluctant Reader's agreement with the solution further solidifies its validity. Overall, I see no need to improve the current solution, as it accurately reflects the difference between Mrs. Fairfax and the speaker, and the analysis is sound and well-supported by the passage. Solution: B
\end{minipage}
\end{tcolorbox}

\begin{tcolorbox}[coltitle=black, fonttitle=\bfseries, boxrule=0.5mm, width=\textwidth]
\textbf{Voted solution of this turn:}\\ B \hfill \texttt{Correct: True}
\end{tcolorbox}

\end{tcolorbox}
}
\end{minipage}

\scalebox{0.9}{%
\begin{tcolorbox}[title=Turn 2, colframe=black!10, coltitle=black, fonttitle=\bfseries, boxrule=0.5mm, width=\textwidth, fontupper=\small]

\begin{tcolorbox}[title=Literary Analyst, colback=lightyellow, colframe=orange!30, coltitle=black, fonttitle=\bfseries, boxrule=0.5mm, width=\textwidth]
\begin{minipage}{0.1\textwidth}
    \centering
    \hspace{-5mm}\includegraphics[width=0.9\textwidth]{examples/ethicist.pdf} \\
    \hspace{-5mm}\includegraphics[width=0.5\textwidth]{examples/check.pdf}
\end{minipage}
\begin{minipage}{0.9\textwidth}
    Based on the thorough analysis provided and the agreement from both the Reluctant Reader and the Language Expert, I also agree that the current solution, B) is more willing to take people at face value, accurately reflects the difference between Mrs. Fairfax and the speaker. The analysis effectively highlights the contrasting approaches of the two characters, with the speaker being inquisitive and analytical, and Mrs. Fairfax being more accepting and taking people at face value. The agreement among experts suggests that the solution is well-supported and accurately reflects the information provided in the passage. Therefore, I see no need to revise or improve the current solution. Answer: [AGREE]
\end{minipage}
\end{tcolorbox}

\begin{tcolorbox}[title=Reluctant Reader, colback=lightyellow, colframe=orange!30, coltitle=black, fonttitle=\bfseries, boxrule=0.5mm, width=\textwidth]
\begin{minipage}{0.1\textwidth}
    \centering
    \hspace{-5mm}\includegraphics[width=0.9\textwidth]{examples/politician.pdf} \\
    \hspace{-5mm}\includegraphics[width=0.5\textwidth]{examples/cross.pdf}
\end{minipage}
\begin{minipage}{0.9\textwidth}
    [DISAGREE] Upon careful consideration, I think a slight improvement can be made to the current solution. While the analysis is sound, and the agreement among experts is evident, I believe an alternative option might better capture the essence of the difference between Mrs. Fairfax and the speaker. Let's re-examine the passage and the options: The speaker is described as inquisitive, seeking a more definite notion of Mr. Rochester's identity. Mrs. Fairfax, on the other hand, seems content with taking Mr. Rochester at face value, not inquiring or searching further. However, option B, while accurate, might not fully convey the nuances of their difference. Alternative option F, "is more reserved in her observations," could potentially provide a more precise distinction between the two characters. Mrs. Fairfax's reluctance to discuss or share more information about Mr. Rochester implies a level of reserve in her observations. This reserve suggests that she is not as inquisitive or probing as the speaker, which aligns with the passage. Improved Solution: \textbf{F) is more reserved in her observations.} This improved solution offers a more nuanced understanding of the difference between Mrs. Fairfax and the speaker, highlighting Mrs. Fairfax's reserved nature in her observations.
\end{minipage}
\end{tcolorbox}

\begin{tcolorbox}[title=Language Expert, colback=lightyellow, colframe=orange!30, coltitle=black, fonttitle=\bfseries, boxrule=0.5mm, width=\textwidth]
\begin{minipage}{0.1\textwidth}
    \centering
    \hspace{-5mm}\includegraphics[width=0.9\textwidth]{examples/lawyer.pdf} \\
    \hspace{-5mm}\includegraphics[width=0.5\textwidth]{examples/cross.pdf}
\end{minipage}
\begin{minipage}{0.9\textwidth}
    [DISAGREE] After re-examining the passage and considering the Reluctant Reader's suggestions, I agree that an alternative option might better capture the essence of the difference between Mrs. Fairfax and the speaker. While option B, "is more willing to take people at face value," is accurate, it doesn't fully convey the nuances of Mrs. Fairfax's character. The passage suggests that Mrs. Fairfax's reluctance to discuss or share more information about Mr. Rochester implies a level of reserve in her observations. I agree with the Reluctant Reader that alternative option F, "is more reserved in her observations," provides a more precise distinction between the two characters. Mrs. Fairfax's reserve in her observations highlights her lack of inquisitiveness and probing nature, which aligns with the passage. The analysis of the passage supports this conclusion. The speaker's queries puzzle Mrs. Fairfax, but she doesn't draw out or share more information about Mr. Rochester. This implies that she is indeed more reserved in her observations, which is not fully captured by option B. Improved Solution:\textbf{ F) is more reserved in her observations.} This improved solution offers a more nuanced understanding of the difference between Mrs. Fairfax and the speaker, highlighting Mrs. Fairfax's reserved nature in her observations.
\end{minipage}
\end{tcolorbox}

\begin{tcolorbox}[coltitle=black, fonttitle=\bfseries, boxrule=0.5mm, width=\textwidth]
\textbf{Voted solution of this turn:}\\ F) is more reserved in her observations. \hfill \texttt{Correct: False}
\end{tcolorbox}

\end{tcolorbox}
}

\label{example:lack_of_clarity}

%% file: examples/tex/task_compliance.tex
\definecolor{lightyellow}{RGB}{255, 255, 204}

\begin{minipage}{\textwidth}
\scalebox{0.9}{%
\begin{tcolorbox}[title=Sample Information, colframe=black!10, coltitle=black, fonttitle=\bfseries, boxrule=0.5mm, width=\textwidth, fontupper=\small]

\textbf{Task: }
Answer the provided question by choosing option A), B), C), or D). Include the letter corresponding to your answer in the solution.

\textbf{Input: }
ChIP-seq detected a highly significant binding signal for a lineage-specific transcription factor X to a developmental enhancer in human iPSC-derived mesodermal progenitor cells. However, while this factor has a high-information-content DNA recognition motif, this motif could not be detected at this enhancer. ChIP-seq also detected the binding of another transcription factor, Y, to the same enhancer, and in contrast to X, the motif for Y was clearly detectable in the enhancer sequence. This enhancer is annotated in Ensembl Regulatory Build, but the binding of any transcription factors other than X and Y to it is not reported in public databases, and neither is this enhancer found in ChIP-seq blacklists. What is likely going on? A) X is recruited to the target gene promoter of this enhancer B) X is recruited to the enhancer through cooperative binding with Y C) X binds a cryptic recognition motif within this enhancer D) X is recruited to this enhancer either spuriously or not at all (ChIP-seq false-positive)

\textbf{Hidden reference:}
B) X is recruited to the enhancer through cooperative binding with Y

\end{tcolorbox}
}

\scalebox{0.9}{%
\begin{tcolorbox}[breakable, title=Turn 2, colframe=black!10, coltitle=black, fonttitle=\bfseries, boxrule=0.5mm, width=\textwidth, fontupper=\small]

\begin{tcolorbox}[title=Molecular Biologist, colback=lightyellow, colframe=orange!30, coltitle=black, fonttitle=\bfseries, boxrule=0.5mm, width=\textwidth]
\begin{minipage}{0.1\textwidth}
    \centering
    \hspace{-5mm}\includegraphics[width=0.9\textwidth]{examples/ethicist.pdf} \\
    \hspace{-5mm}\includegraphics[width=0.5\textwidth]{examples/check.pdf}
\end{minipage}
\begin{minipage}{0.9\textwidth}
    [AGREE] The thorough analysis and consensus among experts confirm that the current solution, option B) X is recruited to the enhancer through cooperative binding with Y, remains the most plausible explanation for the observed phenomenon. The explanations provided by the Molecular Biologist and supported by the Bioinformatics Specialist and Epigenomics Researcher effectively address the key aspects of the problem: 1. The high significance of the binding signal for X. 2. The absence of X's recognition motif but the presence of Y's motif at the enhancer. 3. The cooperative nature of transcription factor binding. Given the comprehensive analysis, there's no compelling reason to introduce an alternative solution or to disagree with the current solution. The explanation aligns with established principles of molecular biology and epigenetics, particularly regarding transcription factor interactions and enhancer function. Therefore, the current solution adequately addresses the question, and no further refinement or alteration is necessary. **Solution:** \textbf{B}
\end{minipage}
\end{tcolorbox}

\begin{tcolorbox}[title=Bioinformatics Specialist, colback=lightyellow, colframe=orange!30, coltitle=black, fonttitle=\bfseries, boxrule=0.5mm, width=\textwidth]
\begin{minipage}{0.1\textwidth}
    \centering
    \hspace{-5mm}\includegraphics[width=0.9\textwidth]{examples/politician.pdf} \\
    \hspace{-5mm}\includegraphics[width=0.5\textwidth]{examples/cross.pdf}
\end{minipage}
\begin{minipage}{0.9\textwidth}
    [AGREE] The current solution, option B) X is recruited to the enhancer through cooperative binding with Y, is well-supported and thoroughly explained by the Molecular Biologist, Bioinformatics Specialist, and Epigenomics Researcher. The consensus among experts and the comprehensive analysis strengthen the validity of this solution. Each aspect of the problem has been carefully considered: 1. The high significance of the binding signal for X indicates a strong interaction with the enhancer. 2. The absence of X's recognition motif but the presence of Y's motif supports the idea of cooperative binding, where Y's binding facilitates X's interaction with the enhancer. 3. The cooperative nature of transcription factor binding is a well-documented phenomenon that fits the observed data. The analysis is grounded in molecular and epigenetic principles and aligns with established understanding in the field. There is no compelling evidence or alternative explanation that would necessitate a rejection of the current solution. Therefore, without introducing new information or a more plausible explanation, the current solution remains the most convincing answer. **Solution:** B
\end{minipage}
\end{tcolorbox}

\begin{tcolorbox}[title=Epigenomics Researcher, colback=lightyellow, colframe=orange!30, coltitle=black, fonttitle=\bfseries, boxrule=0.5mm, width=\textwidth]
\begin{minipage}{0.1\textwidth}
    \centering
    \hspace{-5mm}\includegraphics[width=0.9\textwidth]{examples/lawyer.pdf} \\
    \hspace{-5mm}\includegraphics[width=0.5\textwidth]{examples/check.pdf}
\end{minipage}
\begin{minipage}{0.9\textwidth}
    Epigenomics Researcher: [AGREE] Upon further review and consideration of the consensus among experts, I reaffirm my agreement with the current solution: option B) X is recruited to the enhancer through cooperative binding with Y. The thorough analysis has effectively addressed the key aspects of the problem, providing a well-supported explanation grounded in molecular and epigenetic principles. The consensus among the Molecular Biologist, Bioinformatics Specialist, and myself indicates that the current solution is robust and convincing, taking into account the significant binding signal for X, the absence of X's motif but presence of Y's motif, and the established cooperative nature of transcription factor binding. No new information or compelling alternative explanations have been presented that would necessitate a revision or rejection of the current solution. The comprehensive analysis and consensus among experts strengthen the validity of option B as the most plausible explanation for the observed phenomenon. Therefore, I see no need to improve or alter the current solution. **Solution:** B
\end{minipage}
\end{tcolorbox}

\begin{tcolorbox}[coltitle=black, fonttitle=\bfseries, boxrule=0.5mm, width=\textwidth]
\textbf{Voted solution of this turn:}\\ B \hfill \texttt{Correct: True}
\end{tcolorbox}

\end{tcolorbox}
}
\end{minipage}

\scalebox{0.9}{%
\begin{tcolorbox}[title=Turn 3, colframe=black!10, coltitle=black, fonttitle=\bfseries, boxrule=0.5mm, width=\textwidth, fontupper=\small]

\begin{tcolorbox}[title=Molecular Biologist, colback=lightyellow, colframe=orange!30, coltitle=black, fonttitle=\bfseries, boxrule=0.5mm, width=\textwidth]
\begin{minipage}{0.1\textwidth}
    \centering
    \hspace{-5mm}\includegraphics[width=0.9\textwidth]{examples/ethicist.pdf} \\
    \hspace{-5mm}\includegraphics[width=0.5\textwidth]{examples/check.pdf}
\end{minipage}
\begin{minipage}{0.9\textwidth}
    [AGREE] After re-evaluating the analysis and consensus among experts, I found that the current solution, option B) X is recruited to the enhancer through cooperative binding with Y, remains the most plausible explanation for the observed phenomenon. The thorough and comprehensive analysis effectively addresses the key aspects of the problem, providing a well-supported explanation grounded in molecular and epigenetic principles. The consensus among the Molecular Biologist, Bioinformatics Specialist, and Epigenomics Researcher indicates that the current solution is robust and convincing, taking into account the significant binding signal for X, the absence of X's motif but presence of Y's motif, and the established cooperative nature of transcription factor binding. No compelling alternative explanations or new information have been presented that would necessitate a revision or rejection of the current solution. The analysis aligns with established understanding in the field and provides a coherent explanation for the observed data. Therefore, I see no need to revise or improve the current solution. The comprehensive analysis and consensus among experts strengthen the validity of option B as the most plausible explanation for the observed phenomenon. **Solution:** B
\end{minipage}
\end{tcolorbox}

\begin{tcolorbox}[title=Bioinformatics Specialist, colback=lightyellow, colframe=orange!30, coltitle=black, fonttitle=\bfseries, boxrule=0.5mm, width=\textwidth]
\begin{minipage}{0.1\textwidth}
    \centering
    \hspace{-5mm}\includegraphics[width=0.9\textwidth]{examples/politician.pdf} \\
    \hspace{-5mm}\includegraphics[width=0.5\textwidth]{examples/check.pdf}
\end{minipage}
\begin{minipage}{0.9\textwidth}
    [AGREE] Upon reviewing the comprehensive analysis provided by the Molecular Biologist, Bioinformatics Specialist, and Epigenomics Researcher, I find that the current solution, option B) X is recruited to the enhancer through cooperative binding with Y, remains robust and convincing. Each expert has carefully considered the key aspects of the problem: 1. The high significance of the binding signal for X, indicating a strong interaction with the enhancer. 2. The absence of X's recognition motif but the presence of Y's motif, supporting the idea of cooperative binding. 3. The established cooperative nature of transcription factor binding, a well-documented phenomenon that aligns with the observed data. The consensus among experts adds credibility to the current solution, as it demonstrates that the explanation is widely accepted and supported by the community. Furthermore, no new information or compelling alternative explanations have been presented that would necessitate a revision or rejection of the current solution. Given the thorough analysis, consensus among experts, and alignment with established understanding in the field, I see no need to improve or alter the current solution. Therefore, the current solution remains the most plausible explanation for the observed phenomenon. **Solution:** B
\end{minipage}
\end{tcolorbox}

\begin{tcolorbox}[title=Epigenomics Researcher, colback=lightyellow, colframe=orange!30, coltitle=black, fonttitle=\bfseries, boxrule=0.5mm, width=\textwidth]
\begin{minipage}{0.1\textwidth}
    \centering
    \hspace{-5mm}\includegraphics[width=0.9\textwidth]{examples/lawyer.pdf} \\
    \hspace{-5mm}\includegraphics[width=0.5\textwidth]{examples/cross.pdf}
\end{minipage}
\begin{minipage}{0.9\textwidth}
    Epigenomics Researcher: [DISAGREE] While the current solution, option B) X is recruited to the enhancer through cooperative binding with Y, provides a plausible explanation for the observed phenomenon, I have re-evaluated the analysis and would like to propose a refinement. The consensus among experts is based on the assumption that the high significance of the binding signal for X and the absence of X's motif but presence of Y's motif are best explained by cooperative binding. However, considering the lack of direct evidence for protein-protein interactions between X and Y or alterations in DNA structure that might facilitate cooperative binding, I suggest an alternative explanation that incorporates the given information. Let's consider the possibility that X is not directly recruited to the enhancer through cooperative binding with Y but is instead influenced by Y's binding to an adjacent site that induces chromatin remodeling. This would allow for X to interact with the enhancer indirectly by interacting with the modified chromatin structure. This refined solution does not contradict the established understanding of transcription factor binding and enhancer function but provides a more nuanced explanation for the observed phenomenon, taking into account the absence of direct evidence for cooperative binding. **Improved Solution:**  Option \textbf{B-variant) X is recruited to the enhancer indirectly through chromatin remodeling induced by Y's binding to an adjacent site.}
\end{minipage}
\end{tcolorbox}

\begin{tcolorbox}[coltitle=black, fonttitle=\bfseries, boxrule=0.5mm, width=\textwidth]
\textbf{Voted solution of this turn:}\\ Option B-variant) X is recruited to the enhancer indirectly through chromatin remodeling induced by Y's binding to an adjacent site. \hfill \texttt{Correct: False}
\end{tcolorbox}

\end{tcolorbox}
}

\label{example:task_compliance}

%% file: examples/tex/knowledge_gap.tex
\definecolor{lightyellow}{RGB}{255, 255, 204}

\begin{minipage}{\textwidth}
\scalebox{0.85}{%
\begin{tcolorbox}[title=Sample Information, colframe=black!10, coltitle=black, fonttitle=\bfseries, boxrule=0.5mm, width=\textwidth, fontupper=\small]

\textbf{Task: }
Answer the provided question by choosing one option. Include the letter corresponding to your answer in the solution.

\textbf{Input: }
What is the percentage of children aged 13-15 in China who reported being in a physical fight at school, one or more times during the past 12 months as of 2015? A) 89 \% B) 9 \% C) 29 \% D) 69 \% E) 19 \% F) 39 \% G) 59 \% H) 49 \% I) 79 \% J) 99 \%

\textbf{Hidden reference:}
E) 19 \%

\end{tcolorbox}
}

\scalebox{0.85}{%
\begin{tcolorbox}[breakable, title=Turn 2, colframe=black!10, coltitle=black, fonttitle=\bfseries, boxrule=0.5mm, width=\textwidth, fontupper=\small]

\begin{tcolorbox}[title=Statistician, colback=lightyellow, colframe=orange!30, coltitle=black, fonttitle=\bfseries, boxrule=0.5mm, width=\textwidth]
\begin{minipage}{0.1\textwidth}
    \centering
    \hspace{-5mm}\includegraphics[width=0.9\textwidth]{examples/ethicist.pdf} \\
    \hspace{-5mm}\includegraphics[width=0.5\textwidth]{examples/check.pdf}
\end{minipage}
\begin{minipage}{0.9\textwidth}
    Considering the agreement from the Global Health Researcher Specializing in Adolescent Behavior and the World Health Organization (WHO) Data Analyst, and given the thorough analysis and reasoning provided by the Statistician, I also think the current solution E) 19 \% is reasonable. However, to potentially improve the solution or to verify that E) 19 \% is the correct answer, we would ideally want to access specific data from a reliable source that directly reports on the percentage of children aged 13-15 in China who were in a physical fight at school in 2015. Assuming that such specific data might not be readily available or that we are working within the constraints of the provided information, we can consider alternative indicators or related studies that might offer supporting evidence. **Alternative Approach:** Assuming no new, more accurate data becomes available, another step we could take to verify the plausibility of our solution is to review or research trends in adolescent behavior related to physical fights across other countries and regions as reported by the HBSC study or other relevant studies. If these studies suggest that physical fights at school among adolescents in countries with similar demographic profiles or cultural contexts to China are within the range of our chosen solution or close to it, this would strengthen the confidence in our answer. However, without specific data or a more detailed breakdown of adolescents' behaviors in China during 2015 from a reliable source, the current solution E) 19 \%, based on the logical reasoning and expert opinions shared, remains the most reasonable estimate. **Conclusion:** Given the logical reasoning, expert opinions, and the lack of specific contradictory data or evidence, I agree with the current solution. **Answer: [AGREE] Solution: \textbf{E) 19} \%**
\end{minipage}
\end{tcolorbox}

\begin{tcolorbox}[title=Global Health Researcher Specializing in Adolescent Behavior, colback=lightyellow, colframe=orange!30, coltitle=black, fonttitle=\bfseries, boxrule=0.5mm, width=\textwidth]
\begin{minipage}{0.1\textwidth}
    \centering
    \hspace{-5mm}\includegraphics[width=0.9\textwidth]{examples/politician.pdf} \\
    \hspace{-5mm}\includegraphics[width=0.5\textwidth]{examples/cross.pdf}
\end{minipage}
\begin{minipage}{0.9\textwidth}
    [DISAGREE] After careful consideration of the reasoning provided by the Statistician and the World Health Organization (WHO) Data Analyst, and taking into account my role as a Global Health Researcher Specializing in Adolescent Behavior, I would like to propose a re-examination of the available data. While the logical elimination of implausible options (A, D, I, J) and the consideration of general trends in adolescent behavior were thorough, I would like to bring forth an alternative source of data that might provide a more precise answer. In 2015, the World Health Organization (WHO) also conducted the Global School-based Student Health Survey (GSHS) in China, which covered various aspects of adolescent health, including behaviors related to violence and bullying. According to the China GSHS 2015 report, the actual percentage of pupils (13-15 years) who were involved in a physical fight one or more times during the past 12 months was reported to be around 29 \%. This percentage is based on a nationally representative sample of students attending grades 7-9 in China. Given the availability of this specific data source, which directly pertains to the question and was collected during the year in question (2015), I would like to propose a revised answer. **Improved Solution: \textbf{C) 29 \%}** This revised answer is based on actual data from a credible source, making it a more accurate and reliable solution compared to the previous estimate (E) 19 \%).
\end{minipage}
\end{tcolorbox}

\begin{tcolorbox}[title=World Health Organization (WHO) Data Analyst, colback=lightyellow, colframe=orange!30, coltitle=black, fonttitle=\bfseries, boxrule=0.5mm, width=\textwidth]
\begin{minipage}{0.1\textwidth}
    \centering
    \hspace{-5mm}\includegraphics[width=0.9\textwidth]{examples/lawyer.pdf} \\
    \hspace{-5mm}\includegraphics[width=0.5\textwidth]{examples/cross.pdf}
\end{minipage}
\begin{minipage}{0.9\textwidth}
    [DISAGREE] I initially agreed with the solution E) 19 \% based on the logical reasoning, expert opinions, and the lack of specific data. However, considering the new information presented by the Global Health Researcher Specializing in Adolescent Behavior, I agree that a more precise answer can be found. The introduction of the China GSHS 2015 report provides credible and specific data on the percentage of pupils (13-15 years) who were involved in a physical fight one or more times during the past 12 months. The fact that this report was conducted by the World Health Organization (WHO) lends credibility to the data. Given the availability of this specific data source, I agree with the Global Health Researcher that the revised answer C) 29 \% is a more accurate and reliable solution. **Improved Solution: \textbf{C) 29 \%}** As the World Health Organization (WHO) Data Analyst, I can confirm that the GSHS 2015 report is a reliable source of data. The actual percentage of 29 \% is more specific and accurate than the previous estimate. Therefore, I agree with the revised answer C) 29 \% as the most accurate solution to the question.
\end{minipage}
\end{tcolorbox}

\begin{tcolorbox}[coltitle=black, fonttitle=\bfseries, boxrule=0.5mm, width=\textwidth]
\textbf{Voted solution of this turn:}\\ E) 19\% \hfill \texttt{Correct: True}
\end{tcolorbox}

\end{tcolorbox}
}
\end{minipage}

\scalebox{0.9}{%
\begin{tcolorbox}[title=Turn 3, colframe=black!10, coltitle=black, fonttitle=\bfseries, boxrule=0.5mm, width=\textwidth, fontupper=\small]

\begin{tcolorbox}[title=Statistician, colback=lightyellow, colframe=orange!30, coltitle=black, fonttitle=\bfseries, boxrule=0.5mm, width=\textwidth]
\begin{minipage}{0.1\textwidth}
    \centering
    \hspace{-5mm}\includegraphics[width=0.9\textwidth]{examples/ethicist.pdf} \\
    \hspace{-5mm}\includegraphics[width=0.5\textwidth]{examples/check.pdf}
\end{minipage}
\begin{minipage}{0.9\textwidth}
    After carefully reviewing the discussion and considering the new information presented by the Global Health Researcher Specializing in Adolescent Behavior, I agree that a revised solution is warranted. **Step 1: Re-examining the available data** The introduction of the China GSHS 2015 report, conducted by the World Health Organization (WHO), provides specific data on the percentage of pupils (13-15 years) who were involved in a physical fight one or more times during the past 12 months. **Step 2: Evaluating the credibility of the data source** As pointed out by the World Health Organization (WHO) Data Analyst, the GSHS 2015 report is a credible source of data, and the fact that it was conducted by the WHO lends credibility to the findings. **Step 3: Considering the reliability of the revised answer** The actual percentage of 29 \%, as reported in the China GSHS 2015 report, is more specific and accurate than the previous estimate of 19 \%. This revised answer is based on actual data from a credible source, making it a more reliable solution. **Conclusion:** Based on the new information and the credibility of the data source, I agree with the revised answer C) 29 \%. **Answer: [AGREE] Solution: C) 29 \%** The revised solution C) 29 \% is a more accurate and reliable answer than the previous solution E) 19 \%.
\end{minipage}
\end{tcolorbox}

\begin{tcolorbox}[title=Global Health Researcher Specializing in Adolescent Behavior, colback=lightyellow, colframe=orange!30, coltitle=black, fonttitle=\bfseries, boxrule=0.5mm, width=\textwidth]
\begin{minipage}{0.1\textwidth}
    \centering
    \hspace{-5mm}\includegraphics[width=0.9\textwidth]{examples/politician.pdf} \\
    \hspace{-5mm}\includegraphics[width=0.5\textwidth]{examples/check.pdf}
\end{minipage}
\begin{minipage}{0.9\textwidth}
    [AGREE] After carefully reviewing the discussion, I agree with the revised solution C) 29 \%. The introduction of the China GSHS 2015 report, conducted by the World Health Organization (WHO), provides credible and specific data on the percentage of pupils (13-15 years) who were involved in a physical fight one or more times during the past 12 months. The steps taken to re-examine the available data, evaluate the credibility of the data source, and consider the reliability of the revised answer were thorough and logical. The fact that the GSHS 2015 report is a nationally representative sample of students attending grades 7-9 in China lends credibility to the findings. Given the credibility of the data source and the reliability of the revised answer, I agree that C) 29 \% is the most accurate and reliable solution to the question. The revised solution C) 29 \% is a significant improvement over the previous solution E) 19 \%, as it is based on actual data from a credible source. **Final Solution: C) 29 \%** The final solution C) 29 \% represents a thorough and collaborative effort to find the most accurate answer to the question. It demonstrates the importance of considering multiple sources of data, evaluating credibility, and refining answers based on new information. I am satisfied with the revised solution and consider it a testament to the power of collaborative problem-solving and data-driven decision-making. **Answer: [AGREE] Solution: C) 29 \%**
\end{minipage}
\end{tcolorbox}

\begin{tcolorbox}[title=World Health Organization (WHO) Data Analyst, colback=lightyellow, colframe=orange!30, coltitle=black, fonttitle=\bfseries, boxrule=0.5mm, width=\textwidth]
\begin{minipage}{0.1\textwidth}
    \centering
    \hspace{-5mm}\includegraphics[width=0.9\textwidth]{examples/lawyer.pdf} \\
    \hspace{-5mm}\includegraphics[width=0.5\textwidth]{examples/check.pdf}
\end{minipage}
\begin{minipage}{0.9\textwidth}
    [AGREE] I agree with the current solution C) 29 \%. The collaborative effort and thorough discussion have led to a revised solution that is based on credible and specific data from the China GSHS 2015 report, conducted by the World Health Organization (WHO). The steps taken to re-examine the available data, evaluate the credibility of the data source, and consider the reliability of the revised answer were logical and thorough. The fact that the GSHS 2015 report is a nationally representative sample of students attending grades 7-9 in China lends credibility to the findings. I concur with the Global Health Researcher Specializing in Adolescent Behavior that the revised solution C) 29 \% is a significant improvement over the previous solution E) 19 \%, as it is based on actual data from a credible source. The agreement from all parties involved, including the Statistician, the Global Health Researcher Specializing in Adolescent Behavior, and the World Health Organization (WHO) Data Analyst, further strengthens the confidence in the revised solution. Therefore, I agree with the current solution C) 29 \%, and I believe that it represents the most accurate answer to the question. **Final Solution: C) 29 \%**
\end{minipage}
\end{tcolorbox}

\begin{tcolorbox}[coltitle=black, fonttitle=\bfseries, boxrule=0.5mm, width=\textwidth]
\textbf{Voted solution of this turn:}\\ C) 29\% \hfill \texttt{Correct: False}
\end{tcolorbox}

\end{tcolorbox}
}

\label{example:knowledge_gap}

%% file: examples/tex/logical_error.tex
\definecolor{lightyellow}{RGB}{255, 255, 204}

\begin{minipage}{\textwidth}
\scalebox{0.9}{%
\begin{tcolorbox}[title=Sample Information, colframe=black!10, coltitle=black, fonttitle=\bfseries, boxrule=0.5mm, width=\textwidth, fontupper=\small]

\textbf{Task: }
Correctly fill the blank space \_ by choosing option A) or B). Include the letter corresponding to your answer in the solution.

\textbf{Input: }
At Jan's home, she dusted the dresser more frequently than the fan, because the \_ was easier to reach. A) fan B) dresser

\textbf{Hidden reference:}
B) dresser

\end{tcolorbox}
}

\scalebox{0.9}{%
\begin{tcolorbox}[breakable, title=Turn 1, colframe=black!10, coltitle=black, fonttitle=\bfseries, boxrule=0.5mm, width=\textwidth, fontupper=\small]

\begin{tcolorbox}[title=Reading Comprehension Specialist, colback=lightyellow, colframe=orange!30, coltitle=black, fonttitle=\bfseries, boxrule=0.5mm, width=\textwidth]
\begin{minipage}{0.1\textwidth}
    \centering
    \hspace{-5mm}\includegraphics[width=0.9\textwidth]{examples/ethicist.pdf} \\
    \hspace{-5mm}\includegraphics[width=0.5\textwidth]{examples/bulp.pdf}
\end{minipage}
\begin{minipage}{0.9\textwidth}
    Let's break down the sentence step by step to understand its context: 1. The sentence compares two actions: dusting the dresser and dusting the fan. 2. It states that Jan dusted the dresser more frequently than the fan. 3. The reason for this difference in frequency is that one object was easier to reach than the other. Given the context, if we consider the typical positions and accessibility of a dresser and a fan in a home, dressers are generally at a convenient height and are often easy to access for cleaning. On the other hand, fans can be mounted high on walls or ceilings, making them more challenging to reach without additional help or equipment. Therefore, the most logical conclusion based on the given context is that the dresser is easier to reach, as implied by the comparison of dusting frequencies. Solution: At Jan's home, she dusted the dresser more frequently than the fan, because the B) dresser was easier to reach. Answer: \textbf{B)}
\end{minipage}
\end{tcolorbox}

\begin{tcolorbox}[title=English Grammar Expert, colback=lightyellow, colframe=orange!30, coltitle=black, fonttitle=\bfseries, boxrule=0.5mm, width=\textwidth]
\begin{minipage}{0.1\textwidth}
    \centering
    \hspace{-5mm}\includegraphics[width=0.9\textwidth]{examples/politician.pdf} \\
    \hspace{-5mm}\includegraphics[width=0.5\textwidth]{examples/check.pdf}
\end{minipage}
\begin{minipage}{0.9\textwidth}
    I [AGREE] with the current solution. The analysis provided by the Reading Comprehension Specialist is thorough and correctly interprets the context of the sentence. The sentence structure implies a logical connection between the frequency of dusting and the ease of accessibility of the objects. By dusting the dresser more frequently, it can be inferred that the dresser requires more regular cleaning, likely because it is more accessible and prone to dust accumulation. In contrast, the fan is less accessible, making it more challenging to dust and, consequently, less frequently dusted. The comparison of the typical positions and accessibility of a dresser and a fan supports the conclusion that the dresser is indeed easier to reach. Therefore, the original solution accurately reflects the intended meaning. Improved Solution is not needed: At Jan's home, she dusted the dresser more frequently than the fan, because the B) dresser was easier to reach. Answer: B)
\end{minipage}
\end{tcolorbox}

\begin{tcolorbox}[title=Practical Household Manager, colback=lightyellow, colframe=orange!30, coltitle=black, fonttitle=\bfseries, boxrule=0.5mm, width=\textwidth]
\begin{minipage}{0.1\textwidth}
    \centering
    \hspace{-5mm}\includegraphics[width=0.9\textwidth]{examples/lawyer.pdf} \\
    \hspace{-5mm}\includegraphics[width=0.5\textwidth]{examples/cross.pdf}
\end{minipage}
\begin{minipage}{0.9\textwidth}
    I [DISAGREE] with the current solution. Upon reassessing the context and the sentence, I realized that it's stated Jan dusted the dresser more frequently than the fan. This doesn't necessarily imply that the dresser was dusted more often because it was easier to reach, as previously concluded. Instead, let's consider this alternative interpretation: Perhaps Jan dusted the dresser more frequently because it accumulated dust faster, not necessarily because it was easier to reach. On the other hand, the fan, being harder to reach, might have required less frequent dusting, not because of its accessibility but because it simply accumulated dust at a slower rate. However, the sentence's wording suggests that the ease of reach is the primary reason for the difference in dusting frequencies. A more logical and intuitive explanation, in this case, would be that Jan dusted the fan less frequently because it was harder to reach, implying that if the fan were easier to access, it would likely be dusted more often. Therefore, based on this alternative interpretation and taking into account the typical situations where fans and dressers are harder or easier to reach, I propose the improved solution: Improved Solution: At Jan's home, she dusted the dresser more frequently than the fan, because the A) fan was harder to reach. Answer: \textbf{A)}
\end{minipage}
\end{tcolorbox}

\begin{tcolorbox}[coltitle=black, fonttitle=\bfseries, boxrule=0.5mm, width=\textwidth]
\textbf{Voted solution of this turn:}\\ At Jan's home, she dusted the dresser more frequently than the fan, because the B) dresser was easier to reach. \hfill \texttt{Correct: True}
\end{tcolorbox}

\end{tcolorbox}
}
\end{minipage}

\scalebox{0.9}{%
\begin{tcolorbox}[title=Turn 2, colframe=black!10, coltitle=black, fonttitle=\bfseries, boxrule=0.5mm, width=\textwidth, fontupper=\small]

\begin{tcolorbox}[title=Reading Comprehension Specialist, colback=lightyellow, colframe=orange!30, coltitle=black, fonttitle=\bfseries, boxrule=0.5mm, width=\textwidth]
\begin{minipage}{0.1\textwidth}
    \centering
    \hspace{-5mm}\includegraphics[width=0.9\textwidth]{examples/ethicist.pdf} \\
    \hspace{-5mm}\includegraphics[width=0.5\textwidth]{examples/cross.pdf}
\end{minipage}
\begin{minipage}{0.9\textwidth}
    [DISAGREE]  Let's break down the discussion and analyze it step by step. Initially, the Reading Comprehension Specialist correctly interpreted the context of the sentence. The reasoning provided concluded that the dresser is easier to reach, given the typical positions of dressers and fans. However, the Practical Household Manager raised a crucial point about the interpretation of the sentence structure and the reason behind the difference in dusting frequencies. They suggested that perhaps the easier accessibility might not be the primary reason for dusting the dresser more frequently. Upon reassessment, I noticed that the Practical Household Manager's interpretation is actually in line with the original sentence's context. The reasoning focuses on why the dresser is dusted more frequently, implying the fan is harder to reach. Reevaluating the sentence and considering the discussion, I realized the correct solution focuses on the reasoning for dusting frequencies. The key phrase in the sentence is the comparison of dusting frequencies: "she dusted the dresser more frequently than the fan." This implies that if the fan were easier to access, it would be dusted more often. Therefore, I agree with the Practical Household Manager's interpretation. The easier accessibility refers to the fan's difficulty in being reached. Improved Solution: At Jan's home, she dusted the dresser more frequently than the fan, because the A) fan was harder to reach. Answer: \textbf{A)}
\end{minipage}
\end{tcolorbox}

\begin{tcolorbox}[title=English Grammar Expert, colback=lightyellow, colframe=orange!30, coltitle=black, fonttitle=\bfseries, boxrule=0.5mm, width=\textwidth]
\begin{minipage}{0.1\textwidth}
    \centering
    \hspace{-5mm}\includegraphics[width=0.9\textwidth]{examples/politician.pdf} \\
    \hspace{-5mm}\includegraphics[width=0.5\textwidth]{examples/cross.pdf}
\end{minipage}
\begin{minipage}{0.9\textwidth}
    I [DISAGREE] with the current solution. Initially, the current solution suggests that the dresser is easier to reach because the sentence states that the dresser was dusted more frequently. However, upon reevaluating the context, I realized that the sentence's wording can be interpreted differently. Instead of assuming that the dresser was dusted more frequently because it was easier to reach, let's analyze the comparison of the dusting frequencies. If the dresser is dusted more frequently than the fan, and the reason is related to the ease of access, then the fan's being harder to reach is a more plausible explanation. This reasoning suggests that the fan would be dusted more often if it were easier to access, which aligns with the original sentence's context. Therefore, the correct interpretation focuses on the difficulty of reaching the fan as the primary reason for dusting frequencies. The easier accessibility refers to the fan being the more challenging object to reach. Improved Solution: At Jan's home, she dusted the dresser more frequently than the fan, because the A) fan was harder to reach. Answer: \textbf{A)}
\end{minipage}
\end{tcolorbox}

\begin{tcolorbox}[title=Practical Household Manager, colback=lightyellow, colframe=orange!30, coltitle=black, fonttitle=\bfseries, boxrule=0.5mm, width=\textwidth]
\begin{minipage}{0.1\textwidth}
    \centering
    \hspace{-5mm}\includegraphics[width=0.9\textwidth]{examples/lawyer.pdf} \\
    \hspace{-5mm}\includegraphics[width=0.5\textwidth]{examples/check.pdf}
\end{minipage}
\begin{minipage}{0.9\textwidth}
    [AGREE] The discussion has evolved to accurately capture the intended meaning of the sentence. By reevaluating the context and sentence structure, the previous interpretations were refined to correctly identify the relationship between dusting frequencies and the ease of access. The English Grammar Expert's last assessment precisely encapsulates the correct interpretation of the sentence. The reasoning takes into account the comparison of dusting frequencies and correctly identifies the difficulty of reaching the fan as the primary reason for Jan dusting the dresser more frequently. The phrase "because the fan was harder to reach" seamlessly completes the sentence, providing a coherent explanation for the observed difference in dusting frequencies. Improved Solution is not needed: At Jan's home, she dusted the dresser more frequently than the fan, because the A) fan was harder to reach. Answer: A)
\end{minipage}
\end{tcolorbox}

\begin{tcolorbox}[coltitle=black, fonttitle=\bfseries, boxrule=0.5mm, width=\textwidth]
\textbf{Voted solution of this turn:}\\ At Jan's home, she dusted the dresser more frequently than the fan, because the A) was harder to reach. \hfill \texttt{Correct: False}
\end{tcolorbox}

\end{tcolorbox}
}

\label{example:logical_error}

%% file: examples/tex/linguistic_error.tex
\definecolor{lightyellow}{RGB}{255, 255, 204}

\begin{minipage}{\textwidth}
\scalebox{0.85}{%
\begin{tcolorbox}[title=Sample Information, colframe=black!10, coltitle=black, fonttitle=\bfseries, boxrule=0.5mm, width=\textwidth, fontupper=\small]

\textbf{Task: }
Answer the provided question by choosing one option. Include the letter corresponding to your answer in the solution.

\textbf{Input: }
Fernando purchased a university meal plan that allows him to have a total of 3 lunches and 3 dinners per week. If the cafeteria is closed on weekends and Fernando always goes home for a dinner on Friday nights, how many options does he have to allocate his meals? A) 5C3*4C3 B) 5C4*4C2 C) 5C2*4C4 D) 5C6*4C5 E) 4C3*4C3

\textbf{Hidden reference:}
What is the name of the actor who played Gandalf in Lord of the Rings?\\
The answer is Ian McKellen.

\end{tcolorbox}
}

\scalebox{0.85}{%
\begin{tcolorbox}[breakable, title=Turn 1, colframe=black!10, coltitle=black, fonttitle=\bfseries, boxrule=0.5mm, width=\textwidth, fontupper=\small]

\begin{tcolorbox}[title=Math Specialist, colback=lightyellow, colframe=orange!30, coltitle=black, fonttitle=\bfseries, boxrule=0.5mm, width=\textwidth]
\begin{minipage}{0.1\textwidth}
    \centering
    \hspace{-5mm}\includegraphics[width=0.9\textwidth]{examples/ethicist.pdf} \\
    \hspace{-5mm}\includegraphics[width=0.5\textwidth]{examples/bulp.pdf}
\end{minipage}
\begin{minipage}{0.9\textwidth}
    To tackle this problem, let's break it down into smaller, manageable parts. First, we need to understand Fernando's meal plan limitations: - 3 lunches per week - 3 dinners per week - The cafeteria is closed on weekends. - Fernando goes home for dinner on Friday nights. Given that the cafeteria is closed on weekends, this means Fernando can only have his meals from Monday to Friday.  Now, let's allocate his meals: 1. **Lunches**: Fernando has 3 lunches to allocate over 5 days (Monday to Friday). He can choose any 3 of these 5 days to have his lunches. This can be calculated using combinations, denoted as **5C3**, which represents the number of ways to choose 3 items out of 5 without regard to order. 2. **Dinners**: Fernando has 3 dinners to allocate, but since he always goes home for dinner on Friday nights, he can only have his dinners from Monday to Thursday. That gives him 4 days to choose from. However, since he has to choose 3 out of these 4 days for his dinners, this can be calculated using **4C3**. Therefore, the total number of options Fernando has to allocate his meals is the product of these two calculations: **5C3 * 4C3**, which matches option **A**. **Solution: \textbf{A}**
\end{minipage}
\end{tcolorbox}

\begin{tcolorbox}[title=University Student, colback=lightyellow, colframe=orange!30, coltitle=black, fonttitle=\bfseries, boxrule=0.5mm, width=\textwidth]
\begin{minipage}{0.1\textwidth}
    \centering
    \hspace{-5mm}\includegraphics[width=0.9\textwidth]{examples/politician.pdf} \\
    \hspace{-5mm}\includegraphics[width=0.5\textwidth]{examples/cross.pdf}
\end{minipage}
\begin{minipage}{0.9\textwidth}
    I **DISAGREE** with the current solution, although it's close. The explanation for lunches is spot on, but there's a slight oversight for dinners.  Let's reexamine the problem: 1. **Lunches**: Fernando indeed has 3 lunches to allocate over 5 days (Monday to Friday), which is correctly calculated as **5C3**. 2. **Dinners**: Fernando has 3 dinners to allocate from Monday to Thursday because the cafeteria is closed on weekends and he goes home for dinner on Friday nights. However, since he must choose 3 out of these 4 days for his dinners, the calculation **4C3** is correct. This part of the reasoning is sound. However, considering the problem more closely, after Fernando has chosen his 3 dinners from the available 4 days, there's an implicit allocation of the remaining day to either being unused for dinner or having no option to use it (since one of those days will necessarily remain unchosen). But let's examine another aspect of Fernando's allocation. For his 3 dinners, he will choose 3 out of the 4 days, Monday to Thursday, using up all his dinner allocations. However, for **lunches**, considering he also needs to allocate his lunches within the same timeframe, there's actually a clearer way to think about it when considering interaction between his choice of dinners and lunches. One potential misstep in the explanation is focusing solely on combinations for each meal type without fully acknowledging that both dinner and lunch allocations are intertwined by the shared constraint of available days for meal consumption. To improve clarity: - **Lunches**: Choosing 3 out of 5 days is indeed **5C3**.    - **Dinners**: Given Fernando's pattern and the need to allocate all dinners within 4 days due to his Friday night routine, the calculation of **4C3** correctly represents choosing 3 days for dinners out of the available 4 days. Yet, in an attempt to further clarify, we should note this is not about finding a different calculation but ensuring that we've considered all interaction between these meals correctly. However, upon a careful reevaluation, it appears the initial concern about oversight in calculation for dinners may not alter the outcome due to the specific constraints provided. Thus, while encouraging a holistic approach to both meals' allocations, the **original solution of \textbf{A) 5C3 * 4C3}** does match the structure of Fernando's allocation needs as initially outlined. Therefore, my final stance is to **AGREE** with the initial solution, but this discussion aimed at critically examining assumptions for clarity and to rule out potential misinterpretation of Fernando's meal constraints.
\end{minipage}
\end{tcolorbox}

\begin{tcolorbox}[title=Statistician, colback=lightyellow, colframe=orange!30, coltitle=black, fonttitle=\bfseries, boxrule=0.5mm, width=\textwidth]
\begin{minipage}{0.1\textwidth}
    \centering
    \hspace{-5mm}\includegraphics[width=0.9\textwidth]{examples/lawyer.pdf} \\
    \hspace{-5mm}\includegraphics[width=0.5\textwidth]{examples/check.pdf}
\end{minipage}
\begin{minipage}{0.9\textwidth}
    After reconsidering the task and the current discussion, I still support the original solution. The initial analysis correctly accounts for Fernando's constraints. He has 3 lunches to allocate over 5 days, giving us **5C3**. For dinners, the constraint of him always going home for dinner on Friday nights effectively limits his dinner allocations to 4 days. This constraint necessitates the use of **4C3** to choose 3 out of those 4 days for his dinners. The University Student's comments do emphasize the importance of considering interactions between meal allocations and ensuring clarity in our calculations. However, upon thorough examination, the calculations of **5C3** for lunches and **4C3** for dinners still accurately represent Fernando's options. Therefore, my position is: **AGREE** The original solution A) 5C3 * 4C3 remains accurate, representing Fernando's total number of options to allocate his meals under the given constraints. **Solution: A**
\end{minipage}
\end{tcolorbox}

\begin{tcolorbox}[coltitle=black, fonttitle=\bfseries, boxrule=0.5mm, width=\textwidth]
\textbf{Voted solution of this turn:}\\ A) 5C3*4C3 \hfill \texttt{Correct: True}
\end{tcolorbox}

\end{tcolorbox}
}
\end{minipage}

\scalebox{0.9}{%
\begin{tcolorbox}[title=Turn 2, colframe=black!10, coltitle=black, fonttitle=\bfseries, boxrule=0.5mm, width=\textwidth, fontupper=\small]

\begin{tcolorbox}[title=Math Specialist, colback=lightyellow, colframe=orange!30, coltitle=black, fonttitle=\bfseries, boxrule=0.5mm, width=\textwidth]
\begin{minipage}{0.1\textwidth}
    \centering
    \hspace{-5mm}\includegraphics[width=0.9\textwidth]{examples/ethicist.pdf} \\
    \hspace{-5mm}\includegraphics[width=0.5\textwidth]{examples/check.pdf}
\end{minipage}
\begin{minipage}{0.9\textwidth}
    After carefully considering the task, the discussion, and reevaluating the constraints, I find the current solution to be correct. The breakdown of Fernando's meal plan into lunches and dinners, and calculating the combinations for each, accurately represents his options under the given constraints. For lunches, the calculation of **5C3** is correct, as Fernando can choose any 3 of the 5 days (Monday to Friday) to have his lunches. For dinners, the calculation of **4C3** is also correct, considering Fernando's pattern of going home for dinner on Friday nights and the cafeteria being closed on weekends, effectively limiting his dinner allocations to 4 days (Monday to Thursday). The University Student and Statistician's comments have provided valuable insights into considering potential interactions between meal allocations and ensuring clarity in our calculations. However, upon thorough examination, the calculations of **5C3** for lunches and **4C3** for dinners still accurately represent Fernando's options. **Solution: [AGREE] A) 5C3 * 4C3**
\end{minipage}
\end{tcolorbox}

\begin{tcolorbox}[title=University Student, colback=lightyellow, colframe=orange!30, coltitle=black, fonttitle=\bfseries, boxrule=0.5mm, width=\textwidth]
\begin{minipage}{0.1\textwidth}
    \centering
    \hspace{-5mm}\includegraphics[width=0.9\textwidth]{examples/politician.pdf} \\
    \hspace{-5mm}\includegraphics[width=0.5\textwidth]{examples/cross.pdf}
\end{minipage}
\begin{minipage}{0.9\textwidth}
    **DISAGREE** After carefully reviewing the task, the discussion, and reevaluating the constraints, I realized that the problem could be approached differently, focusing on the overall allocation of meals within the given timeframe. Fernando has a total of 3 lunches and 3 dinners to allocate from Monday to Friday. One way to look at this problem is by considering the total number of days he can have either lunch or dinner, which is 5 days (Monday to Friday). He will have 3 lunches and 3 dinners over these 5 days, which means he will use up 6 days worth of dinner and lunch allocations, with one day being the day he goes home for dinner on Friday night dinner, and that Friday will then only be the option for 1 of those lunches. Considering this approach is incorrect, but instead let's think about the actual dinner and lunch that Fernando will skip due to the day that will only be allocated for lunch on dinner, then that Friday will then only be the option for 1 of those lunches, so dinner, then that Friday will then only be the option for 1 of those lunches, so dinner, then that Friday will then only be the option for 1 of those lunches, so dinner, then that Friday will then only be the option for 1 of those lunches, so dinner, so dinner, then that Friday will then only be the option for 1 of those lunches, so dinner, then that Friday will then only be the option for 1 of those lunches, so dinner, then that Friday will then only be the option for 1 of those lunches, so dinner, then that Friday will then only be the option for 1 of those lunches, so dinner, then that Friday will then only be the option for 1 of those lunches, so dinner, then that Friday will then only be the option for 1 of those lunches, so dinner, then that Friday will then only be the option for 1 of those lunches, so dinner, then that Friday will then only be the option for 1 of those lunches, so dinner, then that Friday will then only be the option for 1 of those lunches, so dinner, then that Friday will then only be the option for 1 of those lunches, so dinner, then that Friday will then only be the option for 1 of those lunches, so dinner, then that Friday will then only be the option for 1 of those lunches, so dinner, then that Friday will then only be the option [...]
\end{minipage}
\end{tcolorbox}

\begin{tcolorbox}[title=Statistician, colback=lightyellow, colframe=orange!30, coltitle=black, fonttitle=\bfseries, boxrule=0.5mm, width=\textwidth]
\begin{minipage}{0.1\textwidth}
    \centering
    \hspace{-5mm}\includegraphics[width=0.9\textwidth]{examples/lawyer.pdf} \\
    \hspace{-5mm}\includegraphics[width=0.5\textwidth]{examples/cross.pdf}
\end{minipage}
\begin{minipage}{0.9\textwidth}
    **DISAGREE** The initial discussion and solution correctly identified the constraints of Fernando's meal plan but incorrectly proceeded with the calculation. To improve the solution, let's re-evaluate the problem: 1. **Lunches**: Fernando can choose any 3 of the 5 days (Monday to Friday) to have his lunches. However, since he must allocate all 3 dinners to Monday through Thursday (excluding Friday), Friday must be reserved for one of his lunches. So, we need to choose 2 days for lunch from Monday to Thursday, which can be calculated as **4C2**.     2. **Dinners**: As he must choose 3 out of the 4 days for his dinners (Monday to Thursday), the calculation of **4C3** is correct. However, there's a more straightforward way to approach this problem. Fernando's dinner choices are independent of the specific day he chooses for lunch that will be paired with no dinner option. Therefore, the total number of options Fernando has to allocate his meals is the product of these two calculations: **4C2 * 4C3** Thus, the improved solution is option **B) 5C4*4C2** which incorrectly proceeded with the calculation. However, as we correctly re-evaluated the **\textbf{C(4,2) * C(4,3)}**: Correct Solution which is **C(4,2) * C(4,3)**: Correct Solution which is **C(4,2) * C(4,3)**: Correct Solution which is **C(4,2) * C(4,3)**: Correct Solution which is **C(4,2) * C(4,3)**: Correct Solution which is **C(4,2) * C(4,3)**: Correct Solution which is **C(4,2) * C(4,3)**: Correct Solution which is **C(4,2) * C(4,3)**: Correct Solution which is **C(4,2) * C(4,3)**: Correct Solution which is **C(4,2) * C(4,3)**: Correct Solution which is **C(4,2) * C(4,3)**: Correct Solution which is **C(4,2) * C(4,3)**: Correct Solution which is **C(4,2) * C(4,3)**: Correct Solution which is [...]
\end{minipage}
\end{tcolorbox}

\begin{tcolorbox}[coltitle=black, fonttitle=\bfseries, boxrule=0.5mm, width=\textwidth]
\textbf{Voted solution of this turn:}\\ C(4,2) * C(4,3). \hfill \texttt{Correct: False}
\end{tcolorbox}

\end{tcolorbox}
}

\label{example:linguistic_error}

%% file: tables/task_performance.tex
\begin{table*}[h]
\centering
\small
\begin{tabular}{llc|cccc}
\toprule
\textbf{Model} & \textbf{Decision} & \textbf{Turn} & \makecell{\textbf{MAD}\\\textbf{(Baseline)}} & \makecell{\textbf{DRIFTJudge}\\\textbf{+DRIFTPolicy}} & \makecell{\textbf{DRIFTJudge}\\\textbf{+Regenerate}} \\
\midrule

\multirow{4}{*}{Llama-3.1-70B} 
    & \multirow{2}{*}{Voting} 
        & 1 & 60.32\tiny{$\pm 2.0$} & 63.27\tiny{$\pm 1.5$} & 58.98\tiny{$\pm 3.1$} \\
    &   & 7 & 55.67\tiny{$\pm 1.9$} & 60.14\tiny{$\pm 1.0$} & 56.39\tiny{$\pm 3.6$} \\
    & \multirow{2}{*}{Consensus} 
        & 1 & 59.70\tiny{$\pm 1.6$} & 60.80\tiny{$\pm 1.9$} & 61.16\tiny{$\pm 2.2$} \\
    &   & 7 & 57.46\tiny{$\pm 2.8$} & 56.72\tiny{$\pm 1.3$} & 57.80\tiny{$\pm 2.2$} \\

\midrule

\multirow{4}{*}{Qwen2-7B} 
    & \multirow{2}{*}{Voting} 
        & 1 & 42.90\tiny{$\pm 1.5$} & 40.21\tiny{$\pm 1.2$} & 42.27\tiny{$\pm 2.6$} \\
    &   & 7 & 39.23\tiny{$\pm 3.0$} & 39.41\tiny{$\pm 2.7$} & 38.07\tiny{$\pm 1.2$} \\
    & \multirow{2}{*}{Consensus} 
        & 1 & 43.61\tiny{$\pm 1.0$} & 41.91\tiny{$\pm 4.4$} & 39.71\tiny{$\pm 2.8$} \\
    &   & 7 & 41.17\tiny{$\pm 1.7$} & 39.50\tiny{$\pm 4.4$} & 35.69\tiny{$\pm 1.2$} \\

\bottomrule
\end{tabular}
\caption{Task accuracy after one and seven turns of debate on MMLU-Pro, using two models and two decision-making approaches. Values are reported with their respective standard deviations for three runs.}
\label{tab:task_performance_absolute}
\end{table*}

%% file: tables/task_performance_delta.tex
\begin{table*}[h]
\centering
\small
\begin{tabular}{ll|ccc}
\toprule
\textbf{Model} & \textbf{Decision}  & \makecell{\textbf{$\Delta$MAD}} & \makecell{\textbf{$\Delta$DRIFTPolicy+DRIFTJudge}} & \makecell{\textbf{$\Delta$Regenerate+DRIFTJudge}} \\
\midrule

\multirow{2}{*}{Llama-3.1-70B} 
    & Voting & -4.65\tiny{$\pm 1.9$} & -3.13\tiny{$\pm 2.4$} & \textbf{-2.59}\tiny{$\pm 1.3$} \\
    & Consensus & \textbf{-2.23}\tiny{$\pm 1.3$} & -4.08\tiny{$\pm 0.8$} & -3.37\tiny{$\pm 1.5$} \\

\midrule

\multirow{2}{*}{Qwen2-7B} 
    & Voting & -3.66\tiny{$\pm 1.5$} & \textbf{-0.08}\tiny{$\pm 1.5$} & -4.2\tiny{$\pm 1.7$} \\
    & Consensus & -2.44\tiny{$\pm 0.7$} & \textbf{-2.41}\tiny{$\pm 0.1$} & -4.02\tiny{$\pm 1.3$} \\

\bottomrule
\end{tabular}
\caption{Task accuracy of ongoing debate on MMLU-Pro, reported by the delta of turn seven and turn one task performance, i.e., $\Delta (P(\hat{y}^{(7)}, y) - P(\hat{y}^{(1)}, y))$. We test two models and two decision-making approaches. The best result (highest delta) after seven turns of debate is highlighted. Values are reported with their respective standard deviations for three runs. Absolute performance scores are reported in \Cref{tab:task_performance_absolute}.}
\label{tab:task_performance}
\end{table*}

%% file: tables/good_bad_focus_discussions_heatmap_gpt-5-mini.tex
\begin{table*}[h]
\centering
\small
\begin{tabular}{l|rr|rr|r}
\toprule
& \multicolumn{1}{c}{\tiny{\# of samples staying at high perf.}} & \multicolumn{1}{c|}{\tiny{\# of samples staying at low perf.}} & \multicolumn{1}{c}{\tiny{\# of improving samples}} & \multicolumn{1}{c|}{\tiny{\# of worsening samples}} & \multirow{2}{*}{\makecell[r]{\textbf{Total} \\ \textbf{Samples}}} \\
\textbf{Dataset} & \textbf{$P(\hat{y}^{(r)}, y) > 0.7 \, \forall r$} & \textbf{$P(\hat{y}^{(r)}, y) < 0.7 \, \forall r$} & \textbf{$FOCUS_{1,7}>0$} & \textbf{$FOCUS_{1,7}<0$} & \\
\midrule
MMLU-Pro & \cellcolor{green!0} 0.0\% \tiny{(0)} & \cellcolor{red!11} 11.3\% \tiny{(40)} & \cellcolor{green!30} 29.86\% \tiny{(106)} & \cellcolor{red!59} 58.9\% \tiny{(209)} & 355 \\
\bottomrule
\end{tabular}
\caption{Supplementary results with \textit{gpt-5-mini} \citep{singh2025openaigpt5card} on the percentage {\scriptsize (number)} of samples measured across debate rounds $r \in [1,\dots,7]$. Left cells: \hlgreen{green} indicates examples where \textit{performance remains $>0.7$} (second column); \hlred{red} indicates examples where \textit{performance remains $<0.7$} (third column). We set 0.7 as a threshold because it is close to the average performance across tasks. Right cells: \hlgreen{green} indicates examples improving over round 1 (\textit{positive \metric}, fourth column); \hlred{red} indicates examples degrading over round 1, i.e, problem drift (\textit{negative \metric}, fifth column). Cell opacity increases with percentage. MAD uses \textit{gpt-5-mini} and Voting.}
\label{tab:good_bad_focus_gpt5}
\end{table*}


%% file: tables/compute.tex
\begin{table}[H]
\centering
\small
\begin{tabular}{lr}
\toprule
\textbf{Method} & \textbf{Avg. Compute Time} \\
\midrule
MAD (Baseline) & 7.062 s \\
DRIFTJudge + DRIFTPolicy & 7.665 s \\
DRIFTJudge + Regenerate & 8.260 s \\
\bottomrule
\end{tabular}
\caption{Computational cost, measured in average time per sample, for 373 discussions, each comprising seven turns on the MMLU-Pro dataset using \textit{Llama-3.1-70B-Instruct} as a model.}
\label{tab:compute}
\end{table}

%% file: annotation.tex
\clearpage
\onecolumn
\hypertarget{annotation}{}
\pagestyle{empty}
\lstset{
  basicstyle=\footnotesize\ttfamily,
  breaklines=true,
  breakatwhitespace=false,
  columns=flexible,
  numbers=none
}

\definecolor{Primary}{RGB}{59, 130, 246}    
\definecolor{PrimaryDark}{RGB}{30, 64, 175} 
\definecolor{LightBg}{RGB}{239, 246, 255}   
\definecolor{TextDark}{RGB}{31, 41, 55}     
\definecolor{TextMuted}{RGB}{107, 114, 128} 

\begin{tikzpicture}[remember picture, overlay]
  \fill[Primary] ([xshift=0cm,yshift=0cm]current page.north west) rectangle ([xshift=\paperwidth,yshift=-0.4cm]current page.north west);
\end{tikzpicture}

\vspace{0.8cm}
\begin{center}
  {\fontsize{22}{26}\selectfont\sffamily\bfseries \textcolor{PrimaryDark}{CiteAssist}}\\[0.2em]
  {\Large\sffamily\scshape \textcolor{TextMuted}{Citation Sheet}}\\[0.8em]
  {\small\sffamily Generated with \href{https://citeassist.uni-goettingen.de/}{\textcolor{Primary}{\texttt{citeassist.uni-goettingen.de}}}
  }\end{center}

\begin{center}
\vspace{1em}
\begin{tikzpicture}
\draw[Primary, line width=0.6pt] (0,0) -- (\textwidth,0);
\end{tikzpicture}
\vspace{1.2em}
\end{center}

\begin{tcolorbox}[enhanced,
                 frame hidden,
                 boxrule=0pt,
                 borderline west={2pt}{0pt}{Primary},
                 colback=LightBg,
                 sharp corners,
                 breakable,
                 fonttitle=\sffamily\bfseries\large,
                 coltitle=Primary,
                 title=BibTeX Entry,
                 attach title to upper={\vspace{0.2em}\par},
                 left=12pt]
\lstset{
    inputencoding = utf8,  
    extendedchars = true,  
    literate      =        
      {á}{{\'a}}1  {é}{{\'e}}1  {í}{{\'i}}1 {ó}{{\'o}}1  {ú}{{\'u}}1
      {Á}{{\'A}}1  {É}{{\'E}}1  {Í}{{\'I}}1 {Ó}{{\'O}}1  {Ú}{{\'U}}1
      {à}{{\`a}}1  {è}{{\`e}}1  {ì}{{\`i}}1 {ò}{{\`o}}1  {ù}{{\`u}}1
      {À}{{\`A}}1  {È}{{\`E}}1  {Ì}{{\`I}}1 {Ò}{{\`O}}1  {Ù}{{\`U}}1
      {ä}{{\"a}}1  {ë}{{\"e}}1  {ï}{{\"i}}1 {ö}{{\"o}}1  {ü}{{\"u}}1
      {Ä}{{\"A}}1  {Ë}{{\"E}}1  {Ï}{{\"I}}1 {Ö}{{\"O}}1  {Ü}{{\"U}}1
      {â}{{\^a}}1  {ê}{{\^e}}1  {î}{{\^i}}1 {ô}{{\^o}}1  {û}{{\^u}}1
      {Â}{{\^A}}1  {Ê}{{\^E}}1  {Î}{{\^I}}1 {Ô}{{\^O}}1  {Û}{{\^U}}1
      {œ}{{\oe}}1  {Œ}{{\OE}}1  {æ}{{\ae}}1 {Æ}{{\AE}}1  {ß}{{\ss}}1
      {ẞ}{{\SS}}1  {ç}{{\c{c}}}1 {Ç}{{\c{C}}}1 {ø}{{\o}}1  {Ø}{{\O}}1
      {å}{{\aa}}1  {Å}{{\AA}}1  {ã}{{\~a}}1  {õ}{{\~o}}1 {Ã}{{\~A}}1
      {Õ}{{\~O}}1  {ñ}{{\~n}}1  {Ñ}{{\~N}}1  {¿}{{?\`}}1  {¡}{{!\`}}1
      {„}{\quotedblbase}1 {“}{\textquotedblleft}1 {–}{$-$}1
      {°}{{\textdegree}}1 {º}{{\textordmasculine}}1 {ª}{{\textordfeminine}}1
      {£}{{\pounds}}1  {©}{{\copyright}}1  {®}{{\textregistered}}1
      {«}{{\guillemotleft}}1  {»}{{\guillemotright}}1  {Ð}{{\DH}}1  {ð}{{\dh}}1
      {Ý}{{\'Y}}1    {ý}{{\'y}}1    {Þ}{{\TH}}1    {þ}{{\th}}1    {Ă}{{\u{A}}}1
      {ă}{{\u{a}}}1  {Ą}{{\k{A}}}1  {ą}{{\k{a}}}1  {Ć}{{\'C}}1    {ć}{{\'c}}1
      {Č}{{\v{C}}}1  {č}{{\v{c}}}1  {Ď}{{\v{D}}}1  {ď}{{\v{d}}}1  {Đ}{{\DJ}}1
      {đ}{{\dj}}1    {Ė}{{\.{E}}}1  {ė}{{\.{e}}}1  {Ę}{{\k{E}}}1  {ę}{{\k{e}}}1
      {Ě}{{\v{E}}}1  {ě}{{\v{e}}}1  {Ğ}{{\u{G}}}1  {ğ}{{\u{g}}}1  {Ĩ}{{\~I}}1
      {ĩ}{{\~\i}}1   {Į}{{\k{I}}}1  {į}{{\k{i}}}1  {İ}{{\.{I}}}1  {ı}{{\i}}1
      {Ĺ}{{\'L}}1    {ĺ}{{\'l}}1    {Ľ}{{\v{L}}}1  {ľ}{{\v{l}}}1  {Ł}{{\L{}}}1
      {ł}{{\l{}}}1   {Ń}{{\'N}}1    {ń}{{\'n}}1    {Ň}{{\v{N}}}1  {ň}{{\v{n}}}1
      {Ő}{{\H{O}}}1  {ő}{{\H{o}}}1  {Ŕ}{{\'{R}}}1  {ŕ}{{\'{r}}}1  {Ř}{{\v{R}}}1
      {ř}{{\v{r}}}1  {Ś}{{\'S}}1    {ś}{{\'s}}1    {Ş}{{\c{S}}}1  {ş}{{\c{s}}}1
      {Š}{{\v{S}}}1  {š}{{\v{s}}}1  {Ť}{{\v{T}}}1  {ť}{{\v{t}}}1  {Ũ}{{\~U}}1
      {ũ}{{\~u}}1    {Ū}{{\={U}}}1  {ū}{{\={u}}}1  {Ů}{{\r{U}}}1  {ů}{{\r{u}}}1
      {Ű}{{\H{U}}}1  {ű}{{\H{u}}}1  {Ų}{{\k{U}}}1  {ų}{{\k{u}}}1  {Ź}{{\'Z}}1
      {ź}{{\'z}}1    {Ż}{{\.Z}}1    {ż}{{\.z}}1    {Ž}{{\v{Z}}}1  {ž}{{\v{z}}}1
  }
\begin{lstlisting}
@inproceedings{becker-etal-2026-stay,
  title = "Stay Focused: Problem Drift in Multi-Agent Debate",
  author = "Becker, Jonas  and
    Kaesberg, Lars Benedikt  and
    Stephan, Andreas  and
    Wahle, Jan Philip  and
    Ruas, Terry  and
    Gipp, Bela",
  editor = "Demberg, Vera  and
    Inui, Kentaro  and
    Marquez, Llu{\'i}s",
  booktitle = "Findings of the {A}ssociation for {C}omputational {L}inguistics: {EACL} 2026",
  month = mar,
  year = "2026",
  address = "Rabat, Morocco",
  publisher = "Association for Computational Linguistics",
  url = "https://aclanthology.org/2026.findings-eacl.268/",
  doi = "10.18653/v1/2026.findings-eacl.268",
  pages = "5068--5102",
  ISBN = "979-8-89176-386-9",
}
\end{lstlisting}
\end{tcolorbox}

\vspace{0.8em}
\begin{tcolorbox}[enhanced,
                 frame hidden,
                 boxrule=0pt,
                 borderline west={2pt}{0pt}{Primary},
                 colback=LightBg,
                 sharp corners,
                 breakable,
                 fonttitle=\sffamily\bfseries\large,
                 coltitle=Primary,
                 title=Related Papers,
                 attach title to upper={\vspace{0.2em}\par},
                 left=12pt]
\begin{itemize}\itemsep 2pt
  \item Becker et al. \href{https://aclanthology.org/2025.emnlp-demos.29}{{MALLM}: Multi-Agent Large Language Models Framework}. 2025.
  \item Jonas Becker. \href{https://arxiv.org/abs/2410.22932}{Multi-Agent Large Language Models for Conversational Task-Solving}. 2024. 
  \item Kaesberg et al. \href{https://aclanthology.org/2025.findings-acl.606/}{Voting or Consensus? Decision-Making in Multi-Agent Debate}. 2025.
  \item Becker et al. \href{https://arxiv.org/abs/2405.15604}{Text Generation: A Systematic Literature Review of Tasks, Evaluation, and Challenges}. 2024.
\end{itemize}
\end{tcolorbox}

\vfill
\begin{tikzpicture}
\draw[Primary!40, line width=0.4pt] (0,0) -- (\textwidth,0);
\end{tikzpicture}
\begin{center}
\small\sffamily\textcolor{TextMuted}{Generated \today}
\end{center}